%% file: acl_latex.tex
\documentclass[11pt]{article}

\usepackage{acl}

\usepackage{times}
\usepackage{latexsym}

\usepackage[T1]{fontenc}
\usepackage[utf8]{inputenc}

\usepackage{microtype}

\usepackage{inconsolata}

\usepackage{graphicx}

\usepackage{amsmath}
\usepackage{amssymb}
\usepackage{subcaption}
\usepackage{algorithm}
\usepackage{algorithmic}

\usepackage[utf8]{inputenc}
\usepackage{geometry}
\usepackage{booktabs}
\usepackage{multirow}
\usepackage{amssymb} % 用于箭头符号
\usepackage[table,xcdraw]{xcolor} % 用于表格颜色
\usepackage{makecell}

\definecolor{rowblue}{RGB}{218, 232, 252}
\title{Beyond Uniform SVD: Dual-Level Optimization across Columns and Modules for LLM Compression}

\author{
 \textbf{Lin Xu},
 \textbf{Xian Gao},
 \textbf{Ting Liu},
 \textbf{Yuzhuo Fu},
\\
\\
Shanghai Jiao Tong University
\\
}

\begin{document}
\maketitle
\input{components/abstract}

\input{components/introduction}

\input{components/related_work}

\input{components/preliminaries}

\input{components/method}

\input{components/experiments}

\input{components/conclusion}

\clearpage
\input{components/Limitations}
\bibliography{custom}

\clearpage

\appendix

\input{components/appendix}

\end{document}

%% file: components/abstract.tex
\begin{abstract}
Low-rank decomposition, particularly Singular Value Decomposition (SVD), is a pivotal technique for mitigating the storage and computational demands of Large Language Models (LLMs). However, prevalent SVD-based approaches overlook the critical phenomenon that decomposition errors exhibit significant disparity across different components of the parameter matrix, often leading to suboptimal approximation. Furthermore, existing methods lack a direct metric to evaluate the importance of individual weight matrices. To address these limitations, we propose \textbf{Duo-SVD} (\textbf{Du}al-level \textbf{O}ptimization \textbf{SVD}), a novel training-free framework that synergizes optimization at both the column and the module levels. First, Duo-SVD incorporates a \textit{Column-Preserving Strategy} that explicitly retains columns exhibiting high decomposition errors, while applying low-rank approximation solely to those with lower errors. Second, at the module level, we employ a \textit{Module-Adaptive Allocation Strategy} that formulates ratio allocation as a global constrained optimization problem based on perturbation-induced model deviation. Extensive experiments demonstrate that Duo-SVD consistently outperforms state-of-the-art SVD-based baselines and structured pruning methods, establishing it as a superior paradigm for efficient LLM compression.
\end{abstract}

%% file: components/introduction.tex
\section{Introduction}

Large Language Models (LLMs) \citep{model_gpt3,model_llama2, qwen3} have revolutionized Natural Language Processing, yet their escalating parameter scales—ranging from tens to hundreds of billions—pose severe deployment challenges \citep{background_efficient_inference,background_challenge}. To address this, model compression techniques such as Quantization \citep{quantization:optq,quantization_awq}, Pruning \citep{pruner_llm_pruner,laco}, Knowledge Distillation \citep{distillation_minillm,distillation_minillm_resviting}, and Low-Rank Decomposition \citep{slicegpt,layer_sparsity_modegpt} have been widely adopted.

Among these, Singular Value Decomposition (SVD) offers a theoretically grounded approach for compression~\citep{eckart}. However, most existing SVD-based methods \citep{svd_asvd,BasisSharingsvdllm,dobisvd} apply a holistic decomposition strategy. Although approaches like SoLA~\citep{sola} leverage activation sparsity to retain dominant channels, they still lack a deeper exploration into the intrinsic properties of the weight matrices themselves.

In this regard, we identify a critical phenomenon: weight matrices exhibit \textit{structural heterogeneity} concerning SVD decomposition. Specifically, we observe that decomposition errors are non-uniformly distributed: certain columns of the matrix yield significant reconstruction errors, whereas errors in other columns are negligible. This implies that a uniform decomposition strategy, which overlooks such structural heterogeneity, leads to suboptimal compression performance.

\input{figures/Duo-SVD_illustration}

Beyond the internal characteristics of weight matrices, another widely recognized phenomenon is the heterogeneous importance across different modules within LLMs~\citep{layer_sparsity:owl,alphapruning}. This necessitates varying rank retention for different modules. Prior works attempt to solve this via varying heuristics: Bolaco \citep{bolaco} employs Bayesian optimization, SVD-LLM v2 \citep{svdlmv2} relies on truncation error thresholds, and SoLA \citep{sola} minimizes module-wise truncation errors. While effective to a degree, these methods rely on proxy metrics rather than a detailed assessment of each module's impact on the model's actual performance, leaving significant room for optimization.

To bridge these gaps, we propose \textbf{Duo-SVD} (\textbf{Du}al-level \textbf{O}ptimization \textbf{SVD}), a novel training-free framework that synergizes optimization at both the column and module levels. 

First, targeting micro-level \textit{structural heterogeneity}, Duo-SVD incorporates a \textbf{Column-Preserving Strategy}. By leveraging the varying sensitivity of different weight columns, Duo-SVD explicitly retains high-sensitivity columns in their dense form while applying low-rank decomposition to the remaining ones. Furthermore, by exploiting the monotonicity of the decomposition error, it efficiently determines the optimal number of retained columns. Second, regarding macro-level resource distribution, we introduce a \textbf{Module-Adaptive Allocation Strategy}. By evaluating the impact of each module on the model's output under varying compression ratios, we formulate the compression ratio allocation as a global constrained optimization problem. This allows the framework to automatically identify the optimal compression configuration that minimizes the total errors under a strict global parameter budget.

Our key contributions are summarized as follows:
\begin{itemize}
    \item We identify the limitation of uniform SVD in handling the structural heterogeneity of weight matrices and propose a \textit{Column-Preserving Strategy} that integrates exact column retention with low-rank approximation to reduce reconstruction error.
    \item We formulate rank allocation as a global constrained optimization problem, introducing a \textit{Module-Adaptive Allocation Strategy} to precisely manage the compression budget across modules based on perturbation sensitivity.
    \item Extensive experiments demonstrate that Duo-SVD consistently outperforms state-of-the-art SVD baselines and structured pruning methods on both language modeling and downstream tasks, marking a significant advancement in compression efficacy.
\end{itemize}

%% file: figures/Duo-SVD_illustration.tex
\begin{figure*}[t]
    \centering

    \includegraphics[width=\linewidth]{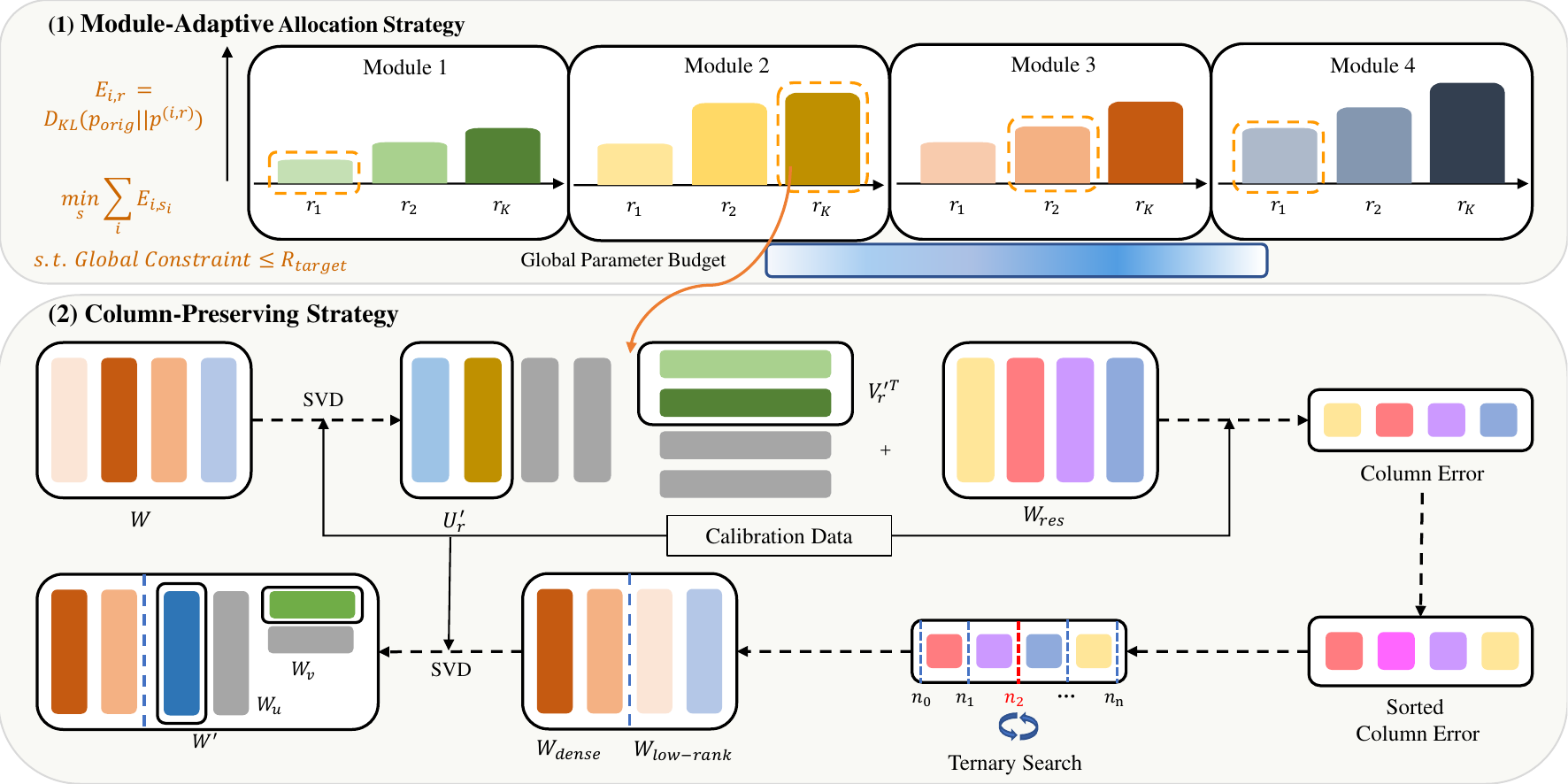}
    
    % 图注
    \caption{Framework of Duo-SVD.
(1) Compute the perturbation error of each module and solve for the module-wise compression ratios that minimize the overall perturbation error. (2) Compute the residual matrices and column-wise errors for each module, then determine which columns to retain and decompose the remaining columns.
}
    
    % 标签，用于文中引用 (如: see Figure \ref{fig:ppl_comparison})
    \label{fig:illustration}
\end{figure*}

%% file: components/related_work.tex
\section{Related Work}

\paragraph{Pruning.}
Pruning methods are categorized into unstructured and structured approaches. Unstructured pruning, such as SparseGPT~\citep{pruner:sparsegpt}, Wanda~\citep{pruner:wanda}, and BESA~\citep{pruner_besa}, operates on individual weights, removing parameters based on importance metrics like Hessian information or activation magnitude. Although effective in preserving model performance, these methods struggle to achieve practical acceleration on GPUs due to irregular sparsity patterns. Conversely, structured pruning reduces model size by removing parameter channels or layers, which is more hardware-friendly. For instance, LLM-Pruner \citep{pruner_llm_pruner} leverages gradient information to remove coupled structures, FLAP \citep{flap} substitutes stable activation patterns with bias terms, and methods like ShortGPT \citep{pruner_shortgpt} and BlockPruner \citep{BlockPruner} identify depth redundancy to prune entire layers.

\paragraph{Low-Rank Decomposition.}
Decomposition techniques can be divided into single-module and module-pair approaches. For single-module decomposition, Standard SVD is typically refined to mitigate accuracy loss: FWSVD \citep{svd_fwsvd} incorporates Fisher information, while ASVD \citep{svd_asvd} scales weight matrices based on input activations. Recognizing that activations often exhibit stronger low-rank characteristics than weights, methods such as AFM \citep{afm}, Bolaco \citep{bolaco}, and Dobi-SVD \citep{dobisvd} perform PCA or its variants on activations to derive column-orthogonal matrices for compression. Other works like SVD-LLM \citep{svdllm} directly analyze the weight-activation product, while Basis Sharing \citep{BasisSharingsvdllm} further improves compression by sharing low-rank matrices across layers. Notably, SoLA \citep{sola} adopts a decoupling strategy to explicitly retain dominant MLP channels. Regarding module-pair decomposition, methods like SliceGPT \citep{slicegpt} and MoDeGPT \citep{layer_sparsity_modegpt} treat two adjacent matrices as a unified block, achieving compression by jointly reducing their intermediate dimension.

%% file: components/preliminaries.tex
\section{Preliminaries}

\paragraph{Singular Value Decomposition.}
For a weight matrix $\mathbf{W} \in \mathbb{R}^{m \times n}$ with $m \le n$ and an input calibration dataset $\mathbf{X} \in \mathbb{R}^{n \times d}$, the compressed representation is formulated as $ \mathbf{W}' = \mathbf{U}'_r {\mathbf{V}'_r}^\top $, where $\mathbf{U}'_r \in \mathbb{R}^{m \times r}$ and ${\mathbf{V}'_r}^\top \in \mathbb{R}^{r \times n}$.
The objective is to minimize the reconstruction error under a rank constraint:
\begin{equation}
\label{eq:svd_obj_final}
\begin{split}
\min_{\mathbf{W}'} \quad & L_{\text{SVD}, r} = \|\mathbf{W} \mathbf{X} - \mathbf{W}' \mathbf{X}\|_F \\
\text{s.t.} \quad & \text{rank}(\mathbf{W}') \le r .
\end{split}
\end{equation}

Following the formulation in \citep{svdllm}, let $\mathbf{H} = \mathbf{X} \mathbf{X}^\top$ and let $\mathbf{S}$ be its Cholesky factor such that $\mathbf{H} = \mathbf{S}\mathbf{S}^\top$.
We apply SVD to the whitened weight matrix $\mathbf{W} \mathbf{S} = \mathbf{U} \mathbf{\Sigma} \mathbf{V}^\top$.
Here, $\mathbf{\Sigma} = \text{diag}(\delta_1, \dots, \delta_m)$ contains singular values sorted in descending order ($\delta_1 \ge \dots \ge \delta_m$).
By retaining the top-$r$ singular values and their corresponding vectors, we obtain the truncated approximation $\mathbf{W} \approx \mathbf{U}_r \mathbf{\Sigma}_r \mathbf{V}_r^\top \mathbf{S}^{-1}$.
This allows us to decompose the original weight into two low-rank factors: $\mathbf{U}'_r = \mathbf{U}_r \sqrt{\mathbf{\Sigma}_r}$ and ${\mathbf{V}'_r}^\top = \sqrt{\mathbf{\Sigma}_r} \mathbf{V}_r^\top \mathbf{S}^{-1}$.
The resulting truncation loss is determined by the discarded singular values:
\begin{equation}
\label{eq:truncation_loss}
L_{\text{SVD}, r}^2 = \sum_{i=r+1}^{m} \delta_i^2.
\end{equation}

\paragraph{Cauchy Interlacing Theorem.}
Let $\mathbf{A} \in \mathbb{R}^{m \times n}$ (where $m \le n$) be a matrix with singular values $\sigma_1(\mathbf{A}) \ge \dots \ge \sigma_m(\mathbf{A})$.
If $\mathbf{B}$ is a submatrix constructed by removing $k$ columns from $\mathbf{A}$, then for every $j \in \{1, \dots, m-k\}$, the singular values of $\mathbf{B}$ satisfy the following interlacing property:
\begin{equation}
\label{eq:cauchy_general}
\sigma_j(\mathbf{A}) \ge \sigma_j(\mathbf{B}) \ge \sigma_{j+k}(\mathbf{A}).
\end{equation}

%% file: components/method.tex
\input{figures/column_errors}

\section{Method}
\label{sec:method}

We present \textbf{Duo-SVD}, a dual-level optimization framework, as shown in Figure~\ref{fig:illustration}. In Section~\ref{sec:column_preserving}, we first introduce the \textit{Column-Preserving Strategy} to address structural heterogeneity within weight matrices. Subsequently, we detail the \textit{Module-Adaptive Allocation Strategy}, which optimizes the global compression budget across different weight matrices, in Section~\ref{sec:module_adaptive}. The complete procedure is summarized in Algorithm~\ref{alg:duo_svd}.

\subsection{Column-Preserving Strategy}
\label{sec:column_preserving}
\paragraph{Motivation: structural Heterogeneity.}
Holistic SVD approaches minimize the global reconstruction error $L_{\text{SVD}} = \|\mathbf{W}\mathbf{X} - \mathbf{W}'\mathbf{X}\|_F$. However, this global objective obscures the \textit{structural heterogeneity} inherent in LLM weight matrices: the approximation error is often highly non-uniform across weight columns. 
To quantify this, let $\mathbf{W}_{\text{res}} = \mathbf{W} - \mathbf{W}'$ denote the residual matrix arising from SVD truncation. We define the truncation error for the $j$-th column, denoted as $L_{\text{col},j}$, by projecting the column-wise residual onto the input activation space:
\begin{equation}
\label{eq:column loss}
L_{\text{col},j} = \|\mathbf{W}_{\text{res}}[:, j] \cdot \mathbf{X}[j, :]\|_F,
\end{equation}
where $\mathbf{W}_{\text{res}}[:, j]$ is the $j$-th column of the residual matrix and $\mathbf{X}[j, :]$ is the corresponding row of the input $\mathbf{X}$.
Empirical observations (as illustrated in Figure~\ref{fig:column errors}) reveal that $\{L_{\text{col},j}\}$ exhibits a structurally concentrated distribution. The truncation error is dominated by a sparse subset of highly sensitive columns that remain consistent across different inputs. Consequently, uniformly compressing these sensitive columns leads to significant accuracy loss.

\paragraph{Hybrid Decomposition and Computation.}
To mitigate structural heterogeneity, we propose a hybrid compression strategy that partitions a weight matrix into a dense component and a low-rank component. 
Let $\mathcal{S} \subset \{1, \dots, n\}$ denote the index set of columns to be preserved exactly, with cardinality $c = |\mathcal{S}|$, and let $\bar{\mathcal{S}}$ denote the complement set for compression.
Accordingly, the weight matrix $\mathbf{W}$ is split into $\mathbf{W}_{\mathcal{S}} \in \mathbb{R}^{m \times c}$ and $\mathbf{W}_{\bar{\mathcal{S}}} \in \mathbb{R}^{m \times (n-c)}$ and we apply low-rank decomposition solely to $\mathbf{W}_{\bar{\mathcal{S}}}$, approximating it as $\mathbf{W}_u \mathbf{W}_v$, where $\mathbf{W}_u \in \mathbb{R}^{m \times r}$ and $\mathbf{W}_v \in \mathbb{R}^{r \times (n-c)}$.

Correspondingly, the forward propagation $\mathbf{Y} = \mathbf{W}' \mathbf{X}$ is computationally decoupled into a dual-pathway formulation. Let $\mathbf{X}_{\mathcal{S}}$ and $\mathbf{X}_{\bar{\mathcal{S}}}$ represent the subsets of input rows corresponding to the preserved and compressed indices, respectively. The output is computed as:
\begin{equation}
\label{eq:hybrid_forward}
\mathbf{W}' \mathbf{X} = \underbrace{\mathbf{W}_{\mathcal{S}} \mathbf{X}_{\mathcal{S}}}_{\text{Dense Path}} + \underbrace{\mathbf{W}_u (\mathbf{W}_v \mathbf{X}_{\bar{\mathcal{S}}})}_{\text{Low-Rank Path}}.
\end{equation}
This formulation ensures that high-sensitivity columns are processed with full precision via the dense path, while the remaining columns are projected through a parameter-efficient bottleneck.

Our goal is to determine the cardinality $c$ and the optimal subset $\mathcal{S}$ that minimize the reconstruction error under a parameter budget $B$. Formally, the optimization problem is:
\begin{equation}
\label{eq:opt_column}
\begin{split}
\min_{c, \mathcal{S}} \quad & \mathcal{F}(c, \mathcal{S}) = \|\mathbf{W}\mathbf{X} - \mathbf{W}'(\mathcal{S}, r)\mathbf{X}\|_F^2 \\
\text{s.t.} \quad & \mathcal{S} \subset \{1, \dots, n\}, \quad |\mathcal{S}| = c, \\
& mc + r(m + n - c) \le B.
\end{split}
\end{equation}
Here, $mc$ represents the cost of preserved columns, and $r(m + n - c)$ represents the cost of the SVD factorization for the remaining columns. Note that under a fixed budget $B$, the rank $r$ is inversely constrained by $c$: preserving more dense columns reduces the budget available for the low-rank path, necessitating a lower rank $r$.

\input{figures/fvsc}

\paragraph{Efficient Solver: Greedy Selection and Unimodality Analysis.}
Solving Eq.~\ref{eq:opt_column} involves a combinatorial search over $\mathcal{S}$ and $c$, which is typically intractable. We decompose this into two manageable sub-problems.

\textit{Step 1: Optimal Subset Selection.}
For a fixed cardinality $c$, we aim to find the optimal subset $\mathcal{S}^*_c$. Adopting a \textit{Column Independence Assumption}, we evaluate column importance independently, ignoring inter-column correlations.
Under this approximation, the total error reduction decouples into a sum of independent column error terms, simplifying the strategy to a deterministic \textit{Greedy Selection}. By sorting the column errors in descending order $L_{\text{col}, (1)} \ge L_{\text{col}, (2)} \ge \dots \ge L_{\text{col}, (n)}$, the optimal subset is simply the top-$c$ columns: $\mathcal{S}^*_c = \{(1), \dots, (c)\}$. This reduces the objective function to a univariate function of $c$, denoted as $\mathcal{F}(c)$.

\textit{Step 2: Optimal Cardinality Search.}
Exact evaluation is computationally intractable as it necessitates a fresh SVD calculation for the complementary matrix at every candidate $c$. Instead, we analyze the discrete gradient $\Delta \mathcal{F}(c)$ over a step size $k$, where $k$ is the minimal increment in preserved columns that reduces the allowable rank by one unit i.e., $r(c+k) = r(c) - 1$.
Let $\mathbf{W}_{c+k}$ and $\mathbf{W}_{c}$ denote the complementary matrices. Given $m \le n-c$, the error difference is derived as:
\begin{equation}
\label{eq:delta_F_analysis}
\begin{split}
&\Delta \mathcal{F}(c) \\
&= L_{\text{SVD}, r(c)-1}^2(\mathbf{W}_{c+k}) - L_{\text{SVD}, r(c)}^2(\mathbf{W}_{c}) \\
&= \sum_{i=r(c)}^{m}\sigma_{i}^2(\mathbf{W}_{c+k}) - \sum_{i=r(c)+1}^{m}\sigma_{i}^2(\mathbf{W}_{c}) \\
&= \sigma_{r(c)}^2(\mathbf{W}_{c+k}) \\
&\quad + \underbrace{\sum_{i=r(c)+1}^{m} \left( \sigma_{i}^2(\mathbf{W}_{c+k}) - \sigma_{i}^2(\mathbf{W}_{c}) \right)}_{\le 0 \text{ (Eq.~\ref{eq:cauchy_general})}} \\
&\approx \underbrace{\sigma_{r(c)}^2(\mathbf{W}_{c+k})}_{\le \delta_{r(c)}^2(\mathbf{W}) \text{ (Eq.~\ref{eq:cauchy_general})}} - \sum_{j=1}^{k} L_{\text{col}, (c+j)}^2 \\
&\le \delta_{r(c)}^2(\mathbf{W}) - \sum_{j=1}^{k} L_{\text{col}, (c+j)}^2.
\end{split}
\end{equation}
The approximation substitutes the spectral energy loss with the loss of the newly preserved columns, i.e., $-\sum L_{\text{col}}^2$. Consequently, the upper bound of $\Delta \mathcal{F}(c)$ can be viewed as a composition of two competing components: a \textit{cost term} $\delta_{r(c)}^2(\mathbf{W})$ induced by the rank reduction, and a \textit{gain term} $\sum_{j=1}^{k} L_{\text{col}, (c+j)}^2$ derived from column preservation.

As $c$ increases, the rank $r(c)$ decreases, shifting the truncation index toward larger singular values, which causes the cost term to monotonically increase. Conversely, the greedy selection strategy ensures that the gain term exhibits a decreasing trend. The superposition of an increasing cost and a decreasing gain implies that the discrete gradient of the derived upper bound is increasing, establishing the convexity of the approximated objective. Although based on approximation, empirical observations confirm that the exact function $\mathcal{F}(c)$ consistently exhibits unimodality (see Figure~\ref{fig:f vesus c} and further verification in Appendix~\ref{sec:appendix empirical verification}). This alignment justifies the use of Ternary Search to efficiently locate $c^*$ with logarithmic complexity. Note that a similar analysis for $m \ge n-c$ is provided in Appendix~\ref{sec:appendix_proof}.

\input{tables/different_models}

\subsection{Module-Adaptive Allocation Strategy}
\label{sec:module_adaptive}

While the Column-Preserving Strategy optimizes within a matrix, LLMs exhibit significant sensitivity variance across different modules. To address this, we propose a perturbation-based allocation scheme. 

\paragraph{Sensitivity Metric Definition.}
Consider a model with $M$ modules $\{\mathcal{M}_1, \dots, \mathcal{M}_M\}$ and parameter counts $\{P_1, \dots, P_M\}$. Let $\mathcal{R} = \{r_1, \dots, r_K\}$ be a discrete set of candidate compression ratios.
We define the sensitivity of module $i$ at a candidate ratio $r \in \mathcal{R}$ by measuring the divergence between the original model output distribution $p_{\text{orig}}$ and the perturbed output $p^{(i,r)}$.

Specifically, $p^{(i,r)}$ represents the output distribution obtained by compressing module $i$ to ratio $r$ while maintaining all other
 modules at the global target rate $R_{target}$. The corresponding sensitivity error is quantified as:
\begin{equation}
\label{eq:module_error}
E_{i,r} = D_{\text{KL}}(p_{\text{orig}} \| p^{(i,r)}), \quad \forall r \in \mathcal{R}.
\end{equation}

\paragraph{Global Optimization via Integer Programming.}
Our objective is to determine the optimal ratio configuration vector $\mathbf{s} = [s_1, \dots, s_M]$, where each $s_i \in \mathcal{R}$ denotes the selected compression ratio for module $i$. The goal is to minimize the aggregate sensitivity error while strictly adhering to the global parameter budget. This is formulated as a Constrained Optimization Problem:
\begin{equation}
\label{eq:opt_module}
\begin{split}
\mathbf{s}^* = \arg\min_{\mathbf{s} \in \mathcal{R}^M} \quad & \sum_{i=1}^M E_{i, s_i} \\
\text{s.t.} \quad & \frac{\sum_{i=1}^M s_i P_i}{\sum_{i=1}^M P_i} \le R_{\text{target}}.
\end{split}
\end{equation}
This formulation is structurally equivalent to the \textit{Multiple-Choice Knapsack Problem} (MCKP). Given that $M$ is finite and the candidate set $\mathcal{R}$ is small, we employ a Dynamic Programming approach to efficiently solve for the globally optimal allocation $\mathbf{s}^*$.

%% file: figures/column_errors.tex
\begin{figure}[htbp]
  \centering
  
  % --- 第一行 ---
  % 容器占当前栏宽的 48%
  \begin{subfigure}[b]{0.48\linewidth} 
    \centering
    % 图片占满容器（即 100% 的 0.48\linewidth）
    \includegraphics[width=\linewidth]{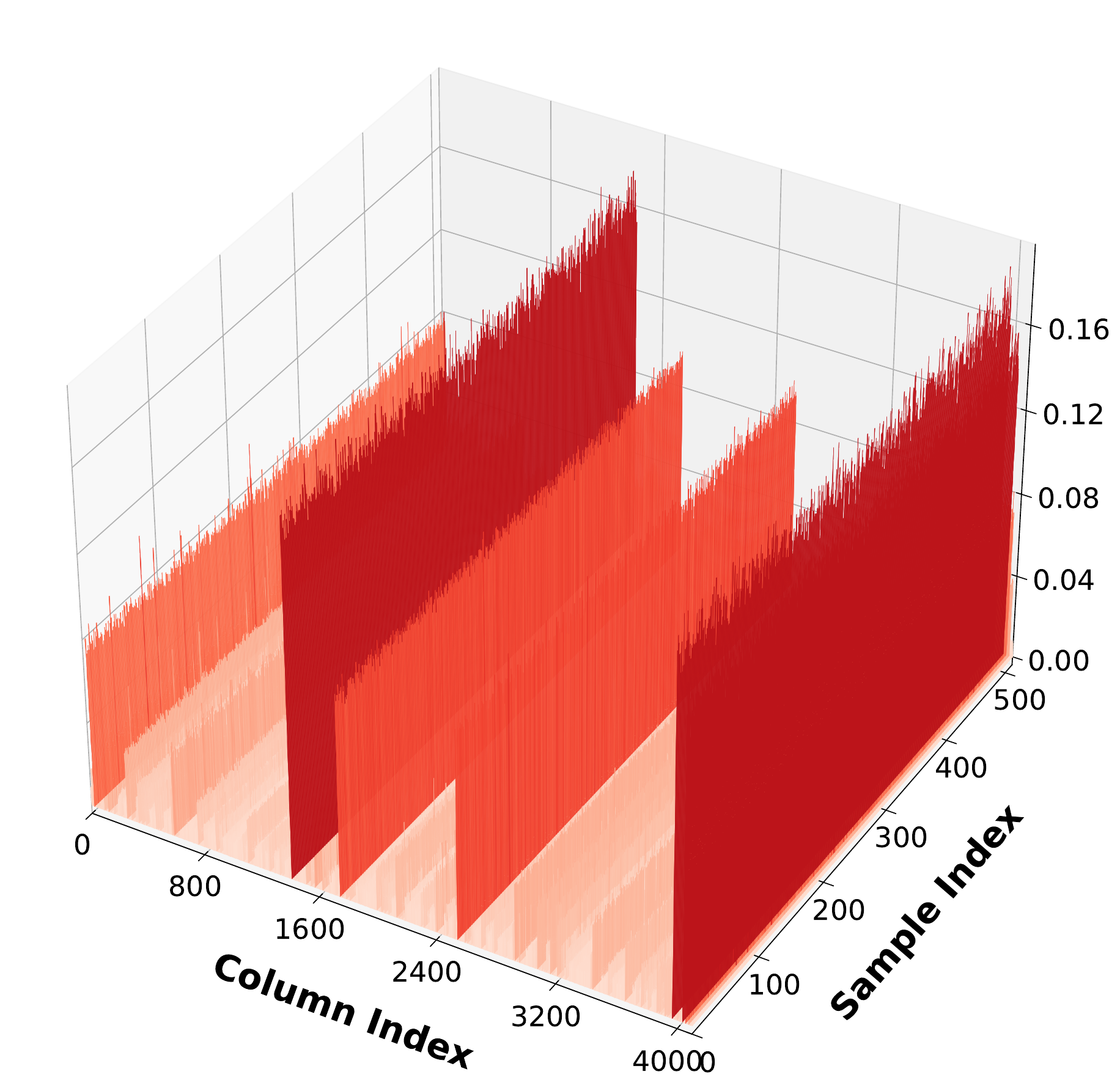}
    \caption{Layer0\_Q\_Proj} % 标题建议简短，避免换行
    \label{fig:sub1}
  \end{subfigure}
  \hfill % 把两张图推向两边
  \begin{subfigure}[b]{0.48\linewidth}
    \centering
    \includegraphics[width=\linewidth]{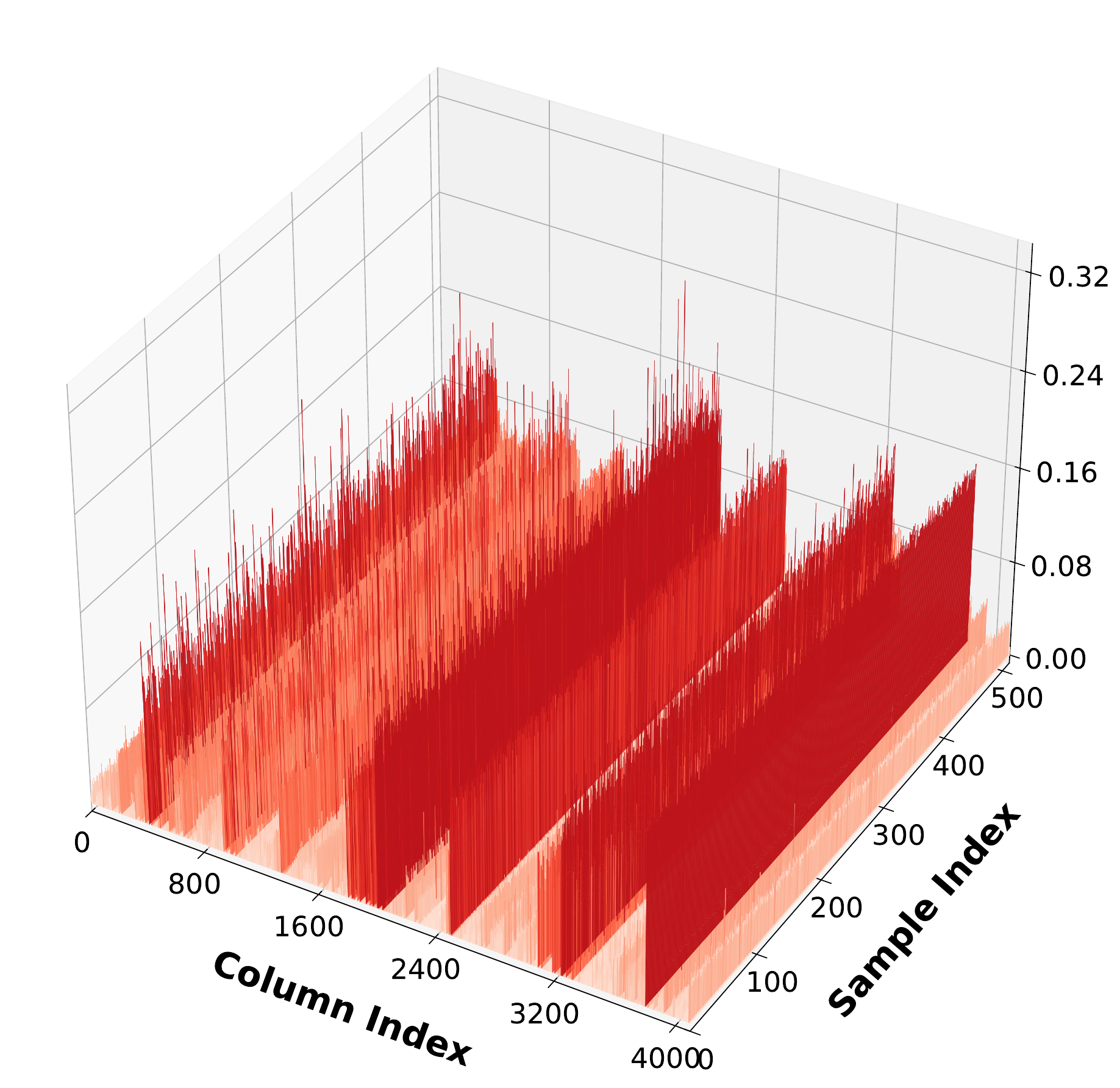}
    \caption{Layer0\_O\_Proj}
    \label{fig:sub2}
  \end{subfigure}
  
  \vspace{0.5em} % 垂直间距可以稍微小一点，节省空间
  
  % --- 第二行 ---
  \begin{subfigure}[b]{0.48\linewidth}
    \centering
    \includegraphics[width=\linewidth]{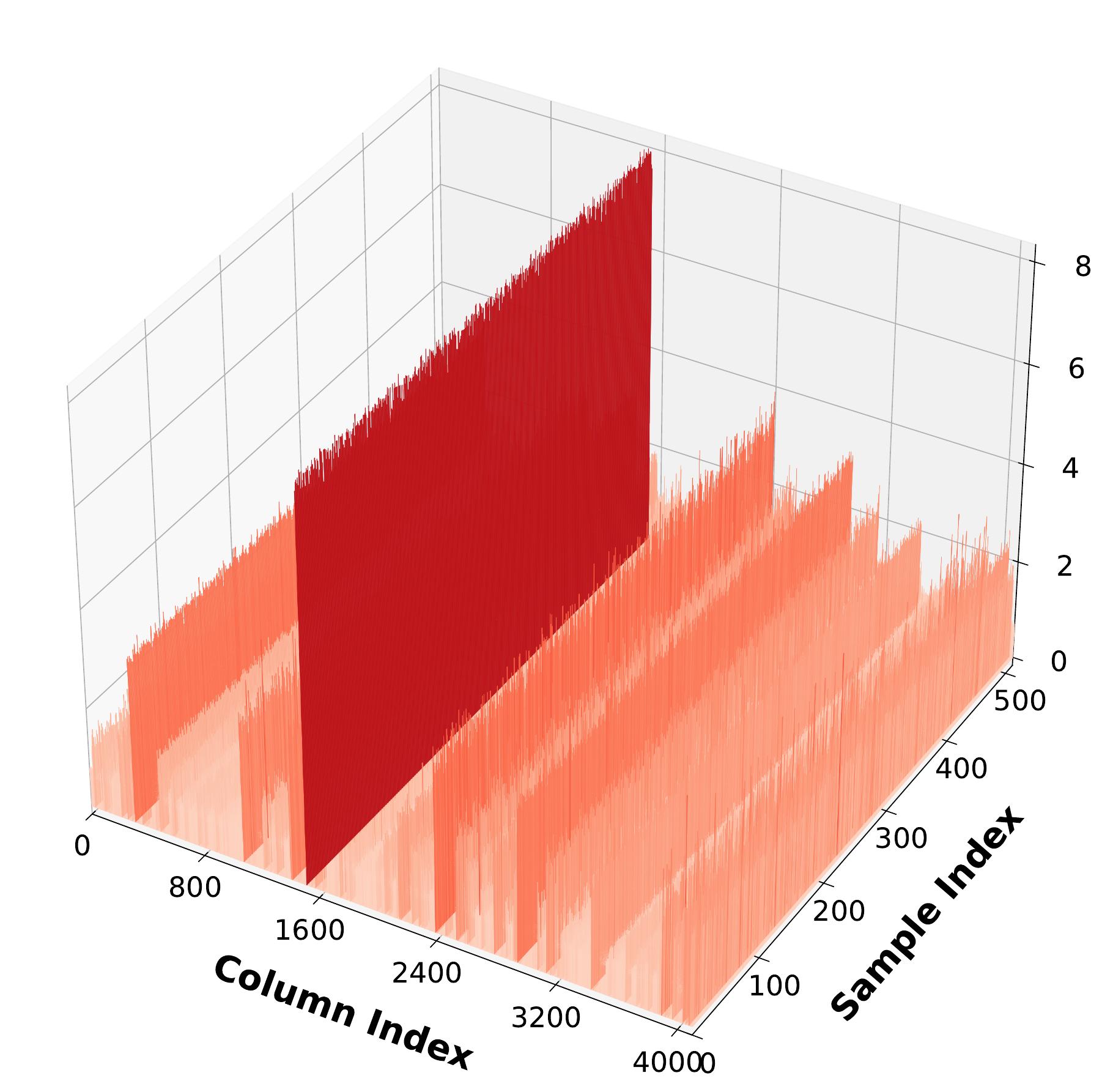}
    \caption{Layer0\_Gate\_Proj}
    \label{fig:sub3}
  \end{subfigure}
  \hfill
  \begin{subfigure}[b]{0.48\linewidth}
    \centering
    \includegraphics[width=\linewidth]{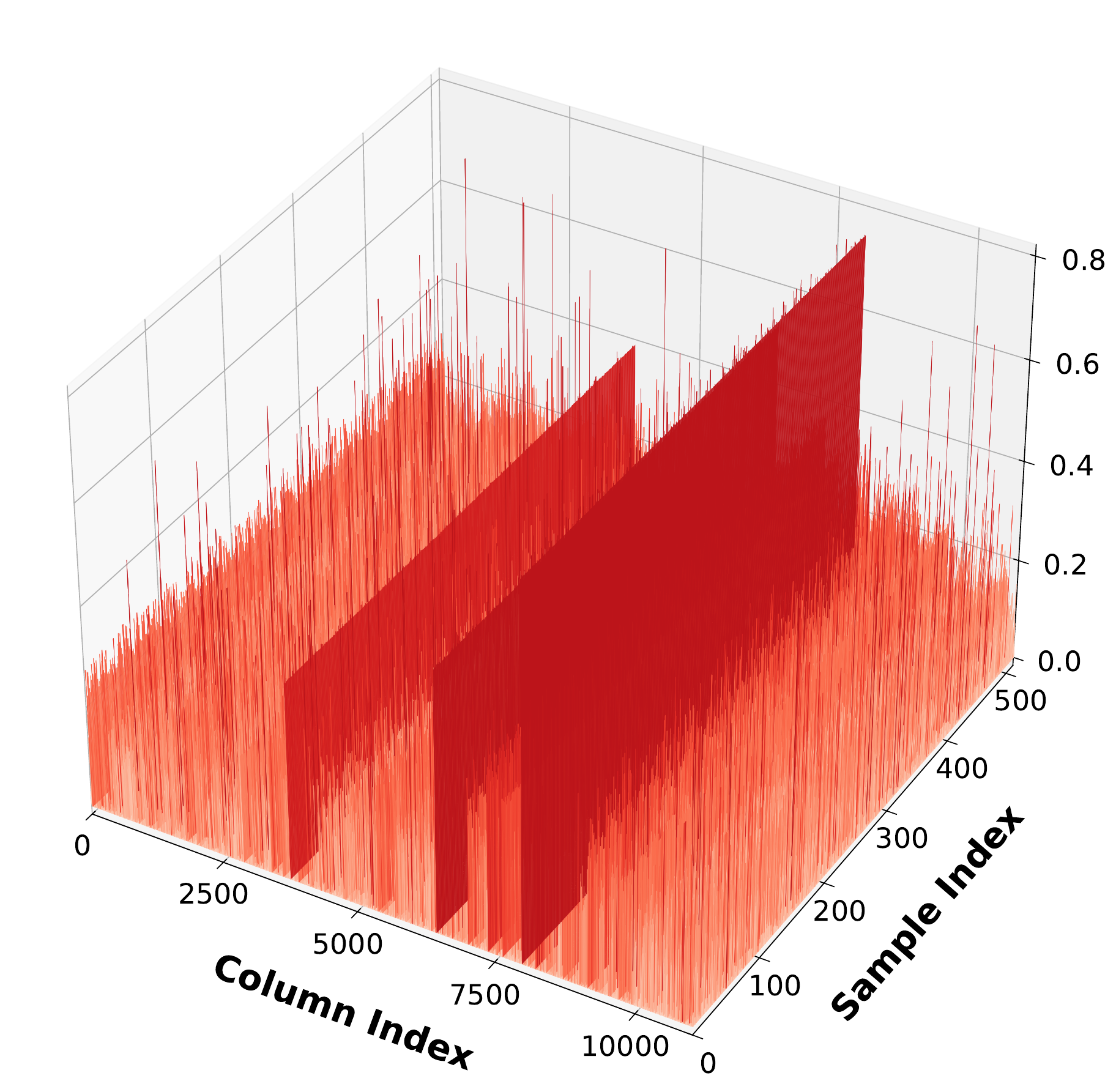}
    \caption{Layer0\_Down\_Proj}
    \label{fig:sub4}
  \end{subfigure}
  
  \caption{Column Errors in LLaMA2-7B}
  \label{fig:column errors}
\end{figure}

%% file: figures/fvsc.tex
\begin{figure}[htbp]
    \centering
    % width=\linewidth 表示图片宽度等于当前列宽
    % file_name 是您保存的图片文件名，扩展名可以省略
    \includegraphics[width=\linewidth]{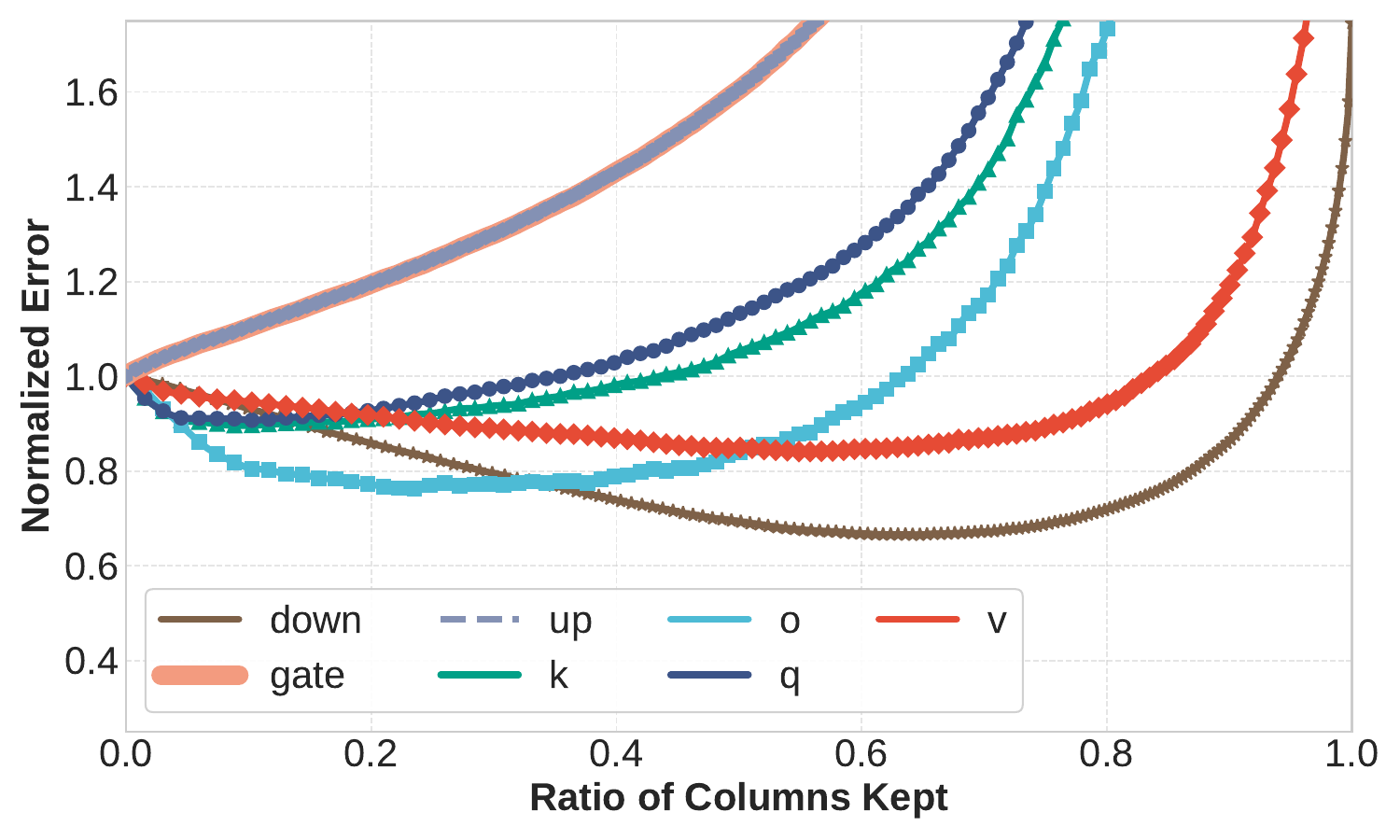}
    
    % 图注
    \caption{$\mathcal{F}(c)$ for LLaMA-2 7B Layer 0 at 20\% Compression Ratio.}
    
    % 标签，用于文中引用 (如: see Figure \ref{fig:ppl_comparison})
    \label{fig:f vesus c}
\end{figure}

%% file: tables/different_models.tex
\begin{table*}[htbp]
\centering
\resizebox{\textwidth}{!}{% 自动调整表格宽度
\begin{tabular}{c|c|c|c|ccccccc}
\hline
\textbf{Model} & \textbf{Method} & \textbf{PPL} $\downarrow$ & \textbf{Avg.} $\uparrow$ & \textbf{MMLU-5shot} & \textbf{PIQA} & \textbf{WinoG.} & \textbf{HellaS.} & \textbf{ARC-e} & \textbf{ARC-c} & \textbf{OBQA} \\ \hline
 \multirow{8}{*}{\textbf{\makecell{LLaMA-2\\7B}}}& Dense & 5.11 & 62.14 & 45.70 & 79.05 & 69.38 & 75.92 & 74.49 & 46.25 & 44.20 \\ \cline{2-11} 
 & LLM-Pruner & 10.55 & 53.89 & 26.20 & 75.95 & 63.38 & 67.83 & 64.31 & 39.93 & 39.60 \\
 & FLAP & 6.76 & 53.07 & 31.90 & 74.54 & 62.98 & 64.74 & 61.28 & 36.43 & 39.60 \\
 & SliceGPT & 8.24 & 46.26 & 26.75 & 64.80 & 62.98 & 49.18 & 55.68 & 31.40 & 33.00 \\
 & SVD-LLM & 7.84 & 45.59 & 26.80 & 65.13 & 62.43 & 51.73 & 47.22 & 27.82 & 38.00 \\
 & Bolaco & 7.31 & 55.23 & 34.30 & 75.09 & 65.61 & 64.33 & 68.19 & 37.48 & 41.60 \\
 & SoLA & 6.52 & 54.33 & 34.10 & 74.65 & \textbf{66.46} & 63.92 & 65.61 & 37.37 & 38.20 \\
\rowcolor{blue!10} \cellcolor{white} & \textbf{Duo-SVD} & \textbf{5.99} & \textbf{57.27} & \textbf{35.41} & \textbf{76.77} & 65.67 & \textbf{70.23} & \textbf{70.50} & \textbf{40.53} & \textbf{41.80} \\ \hline \hline

 \multirow{8}{*}{\textbf{\makecell{LLaMA-2 \\13B}}}& Dense & 4.57 & 65.70 & 55.40 & 80.41 & 72.53 & 79.41 & 77.39 & 49.15 & 45.60 \\ \cline{2-11} 
 & LLM-Pruner & 9.67 & 55.45 & 22.80 & \textbf{77.97} & 60.77 & 71.26 & 67.09 & 44.28 & 44.00 \\
 & FLAP & 5.90 & 57.00 & 41.20 & 75.57 & 67.25 & 69.19 & 65.91 & 39.08 & 40.80 \\
 & SliceGPT & 7.10 & 50.58 & 35.49 & 65.18 & 65.67 & 52.30 & 59.26 & 36.77 & 39.40 \\
 & SVD-LLM & 7.37 & 53.39 & 34.60 & 71.60 & 68.43 & 59.91 & 62.12 & 36.69 & 40.40 \\
 & Bolaco & 6.32 & 59.22 & 43.40 & 76.83 & 65.90 & 69.96 & 70.93 & 42.72 & \textbf{44.80} \\
 & SoLA & 5.61 & 58.83 & 46.10 & 75.57 & 69.77 & 67.35 & 69.15 & 40.70 & 43.20 \\
\rowcolor{blue!10} \cellcolor{white} &  \textbf{Duo-SVD} &  \textbf{5.12} &  \textbf{63.18} &  \textbf{48.73} &  \textbf{77.97} &  \textbf{71.03} &  \textbf{77.40} &  \textbf{75.00} &  \textbf{47.35} &  \textbf{44.80} \\ \hline \hline

\multirow{6}{*}{\textbf{\makecell{Mistral\\-7B}}} & Dense & 4.92 & 68.14 & 62.50 & 82.05 & 73.95 & 81.02 & 79.55 & 53.92 & 44.00 \\ \cline{2-11} 
 & FLAP & 7.11 & 48.29 & 25.90 & 72.31 & 64.09 & 55.94 & 51.05 & 31.91 & 36.80 \\
 & SliceGPT & 9.06 & 43.18 & 25.52 & 59.35 & 61.21 & 45.11 & 51.60 & 30.29 & 29.20 \\
 & SVD-LLM & 9.29 & 50.23 & 25.02 & 70.46 & 64.56 & 58.09 & 69.28 & 37.63 & 26.60 \\
 & SoLA & 6.06 & 56.98 & 44.20 & 73.67 & \textbf{68.75} & 63.32 & \textbf{69.99} & 39.76 & \textbf{39.20} \\
\rowcolor{blue!10} \cellcolor{white} &  \textbf{Duo-SVD} &  \textbf{5.78} &  \textbf{59.37} &  \textbf{47.74} &  \textbf{78.67} &  66.61 &  \textbf{72.20} &  69.15 &  \textbf{43.43} &  37.80 \\ \hline \hline

\multirow{5}{*}{\textbf{\makecell{LLaMA-3.1 \\8B}}} & Dense & 5.84 & 68.36 & 65.32 & 81.18 & 73.64 & 78.89 & 81.10 & 53.41 & 45.00 \\ \cline{2-11} 
 & LLM-Pruner & 15.99 & 50.12 & 28.15 & \textbf{72.69} & 58.09 & 56.75 & 62.50 & 36.86 & 35.80 \\
 & FLAP & 9.23 & 51.52 & 37.85 & 72.14 & 63.69 & 59.04 & 56.31 & 32.00 & 39.60 \\
 & SVD-LLM & 13.72 & 44.46 & 26.51 & 62.57 & 60.77 & 49.93 & 48.02 & 31.23 & 32.20 \\
\rowcolor{blue!10} \cellcolor{white} &  \textbf{Duo-SVD} &  \textbf{8.78} &  \textbf{57.80} &  \textbf{50.55} &  72.63 &  \textbf{70.24} &  \textbf{65.91} &  \textbf{65.61} &  \textbf{38.82} &  \textbf{40.80} \\ \hline
\end{tabular}
}
\caption{Comparison of WikiText-2 perplexity and downstream task accuracy of different methods on LLaMA-2 7B/13B, Mistral-7B, and LLaMA-3.1 8B models under 20\% compression ratio.}
\label{tab:different models}
\end{table*}

%% file: components/experiments.tex
\section{Experiments}

\subsection{Experimental Setup}

\paragraph{Baselines.}
We compare Duo-SVD with structured pruning methods, including LLM-Pruner~\cite{pruner_llm_pruner} and FLAP~\cite{flap}, as well as various decomposition techniques such as SliceGPT~\cite{slicegpt}, Bolaco~\cite{bolaco}, SVD-LLM~\cite{svdllm}, and SoLA~\cite{sola}. Furthermore, we investigate the combination of Duo-SVD with quantization, comparing it against quantization-only baselines. Additional comparisons with BasisSharing~\cite{BasisSharingsvdllm}, Dobi-SVD~\cite{dobisvd}, Laco~\cite{laco}, ShortGPT~\cite{pruner_shortgpt}, and BlockPruner~\cite{BlockPruner} are provided in Appendix~\ref{sec:appendix additional zero shot results}.

\paragraph{Models and Datasets.}
We conduct a comprehensive evaluation on the widely studied LLaMA-2 7B/13B models~\cite{model_llama2}. To further verify the generalization capabilities of our method, we extend our experiments to the Mistral-7B~\cite{model_mistral} and LLaMA-3.1 8B~\cite{llama3} models. We assess model performance across both language modeling and downstream tasks. For language modeling, we utilize Perplexity (PPL) on WikiText-2~\cite{dataset_wiki} datasets as the primary metric. For downstream tasks, we evaluate the model's comprehensive learning capability using 5-shot MMLU~\cite{mmlu} and assess zero-shot reasoning performance across OpenbookQA~\cite{dataset_obqa}, WinoGrande~\cite{dataset_winogrande}, HellaSwag~\cite{dataset_hella}, ARC-e, ARC-c~\cite{dataset_arc}, and PIQA~\cite{dataset_piqa}.

\paragraph{Implementation Details.}
Regarding calibration data, following the protocols of SVD-LLM, we randomly select 256 samples from Wikitext2, each containing 2048 tokens. For the MAAS phase, consistent with Section~\ref{sec:module_adaptive}, the candidate compression ratio set $\mathcal{R}$ consists of 10 elements ranging from 0 to 0.9 (i.e., $\mathcal{R}=\{0, 0.1, \dots, 0.9\}$), and the module-wise sensitivity $E_{i,s}$ is computed using 32 samples, each with 2048 tokens from Wikitext2. All experiments are conducted on NVIDIA RTX 3090 GPUs using the Hugging Face Transformers library, with downstream tasks evaluated via the LM-Evaluation-Harness~\cite{lm_eval_harness}.

\subsection{Experimental Results}

\paragraph{Performance Comparison.}
Table~\ref{tab:different models} summarizes the quantitative results across LLaMA-2 7B/13B, Mistral-7B, and LLaMA-3.1 8B models. As observed, our proposed Duo-SVD consistently outperforms state-of-the-art baselines, achieving the lowest perplexity and the highest average accuracy across all evaluated architectures. For instance, on LLaMA-2 13B, Duo-SVD maintains an impressive average accuracy of 63.18\% with a minimal perplexity of 5.12, effectively preserving the model's generation capabilities. Furthermore, in challenging reasoning benchmarks such as MMLU, our method demonstrates superior robustness, indicating that Duo-SVD effectively retains critical knowledge during compression. Overall, compared to other decomposition-based methods that often suffer from severe performance degradation on diverse model families, Duo-SVD establishes a new state-of-the-art in balancing sparsity and performance.

\input{figures/ppl_across_ratios}
\input{tables/quant}

To further evaluate robustness under aggressive compression, Figure~\ref{fig:ppl across ratio} illustrates the perplexity trends on LLaMA-2 13B across compression ratios from 20\% to 50\%. While baselines such as LLM-Pruner and SliceGPT exhibit rapid performance deterioration as sparsity increases, Duo-SVD demonstrates a significantly flatter growth curve. Even at 50\% ratio, Duo-SVD maintains a stable PPL of 11.24, remarkably lower than SoLA and FLAP, indicating superior preservation of linguistic capabilities.

\paragraph{Combination with Quantization.}
Quantization is a crucial compression method orthogonal to decomposition. To verify the compatibility of Duo-SVD with quantization methods, we applied 3-bit GPTQ~\cite{quantization:optq} quantization to LLaMA-2 7B and LLaMA-3.1 8B compressed by 20\% using Duo-SVD, and compared them against pure quantization methods with similar compression ratios. The results in Table~\ref{tab:quant} indicate that, under comparable compression ratios, the combination of Duo-SVD and quantization achieves lower perplexity, demonstrating excellent compatibility.

\input{figures/module_layer_ratio}

\paragraph{Results of the Module-Adaptive Allocation Strategy.}
Figure~\ref{fig:module types} presents the average compression ratios for different module types in the LLaMA2-13B model. It is evident that there are significant disparities in compression ratios across module types. Specifically, the \texttt{down\_proj} modules consistently maintain a compression ratio lower than the overall target, whereas the \texttt{q\_proj} and \texttt{k\_proj} modules undergo substantial parameter reduction. Additional model results and the comparison with other methods are presented in Appendix~\ref{sec:appendix maas} and Appendix~\ref{sec: appendix different allocation strategy}, respectively.

Furthermore, Figure~\ref{fig:ratio types} illustrates distinct variations in compression ratios across different layers. In general, layers in the first half of the model exhibit lower compression ratios, while those in the second half demonstrate higher ratios, with the first and last layers being notable exceptions. These results provide new insights into the varying importance of different layers under SVD decomposition.

\input{tables/inference}

\paragraph{Inference Efficiency.}
Based on the analysis in Appendix~\ref{sec:appendix efficiency analysis}, a model compressed with a ratio of $r$ theoretically reduces both model size and GEMM computational cost. Here, we further validate the practical performance of Duo-SVD in real-world deployment scenarios. Table~\ref{tab:inference} presents the prefilling latency (Time to First Token, TTFT), autoregressive decoding throughput, and peak memory usage of the LLaMA-2 7B model. In the prefilling phase, Duo-SVD reduces inference latency by lowering computational overhead, while in the autoregressive decoding phase, it improves throughput by reducing parameter count and memory access. The results demonstrate that as the compression ratio increases, Duo-SVD not only significantly reduces memory footprint but also accelerates both the prefilling and decoding stages, highlighting its promising potential for efficient deployment.

\subsection{Ablation Study}
\input{figures/maas_ablation}
\input{figures/calib_datasets}

\paragraph{Effectiveness of Module-Adaptive Allocation Strategy.}
To evaluate the effectiveness of our proposed Module-Adaptive Allocation Strategy (MAAS), we compare the performance of using the Column-Preserving Strategy (CPS) alone against the combination of CPS and MAAS. As shown in Figure~\ref{fig:maas ablation}, compared to using CPS alone, the inclusion of MAAS results in an average reduction in perplexity of approximately 20\% and an average improvement of about 8.5 percentage points in accuracy. These results underscore the significant variance in the importance of different modules within LLMs. Consequently, leveraging these disparities to guide the compression process is crucial for enhancing the performance of compressed models.

\paragraph{Dataset Robustness.}
Currently, we utilize the WikiText-2 dataset for CPS and MAAS. To investigate the impact of different datasets on Duo-SVD, we replaced WikiText-2 with the C4~\citep{dataset_c4} and Alpaca~\citep{alpaca} datasets and re-executed the compression pipeline. Figure~\ref{fig:calibration datasets} illustrates the perplexity on WikiText-2 and C4, as well as the average accuracy on downstream tasks across different calibration datasets. The calibration dataset consistently achieves better performance on its corresponding test metric. Specifically, calibrating with WikiText-2 reduces the perplexity on WikiText-2 by 0.9 compared to calibrating with C4, whereas the perplexity on C4 increases by 0.63. Regarding the average accuracy on downstream tasks, the variation caused by different datasets remains within 1\%. This indicates that switching calibration datasets has a minimal impact on the performance of Duo-SVD in practical tasks, demonstrating the robustness of our method.

%% file: figures/ppl_across_ratios.tex
\begin{figure}[htbp]
    \centering
    % width=\linewidth 表示图片宽度等于当前列宽
    % file_name 是您保存的图片文件名，扩展名可以省略
    \includegraphics[width=\linewidth]{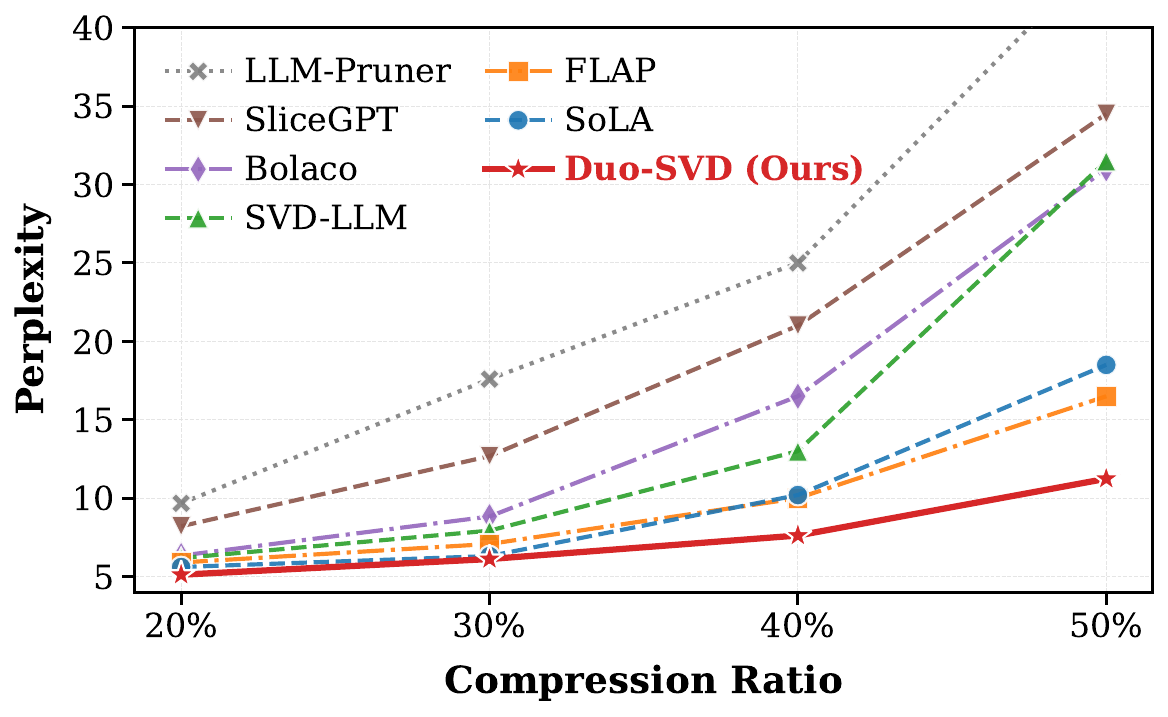}
    
    % 图注
    \caption{Comparison of WikiText-2 perplexity across different compression ratios on LLaMA-2 13B.}
    
    % 标签，用于文中引用 (如: see Figure \ref{fig:ppl_comparison})
    \label{fig:ppl across ratio}
\end{figure}

%% file: tables/quant.tex
\begin{table}[htbp]
\centering
% 确保导言区引用了 \usepackage[table]{xcolor} 和 \usepackage{makecell}
\resizebox{\linewidth}{!}{
    \begin{tabular}{c|c|c|c}
    \toprule
    \textbf{Model} & \textbf{Method} & \textbf{Ratio} & \textbf{PPL}$\downarrow$ \\ 
    \midrule
    
    % --- LLaMA-2 7B Block ---
    % 第1行
     & Dense & -- & 5.12 \\ \cline{2-4}
    % 第2行
     & GPTQ 3-bit & 81.25\% & 8.05 \\ 
    % 第3行：取消整行上色，改为给后3列单独上色
    \multirow{-3}{*}{\textbf{\makecell[c]{LLaMA-2\\7B}}} & 
    \cellcolor{blue!10}\makecell[l]{Duo-SVD +\\GPTQ 4-bit} & 
    \cellcolor{blue!10}80\% & 
    \cellcolor{blue!10}\textbf{6.97} \\ 
    \midrule
    
    % --- LLaMA-3 8B Block ---
    % 第1行
     & Dense & -- & 5.84 \\ \cline{2-4}
    % 第2行
     & GPTQ 3-bit & 81.25\% & 19.28 \\ 
    % 第3行：同样只给后3列上色
    \multirow{-3}{*}{\textbf{\makecell[c]{LLaMA-3.1\\8B}}} & 
    \cellcolor{blue!10}\makecell[l]{Duo-SVD +\\GPTQ 4-bit} & 
    \cellcolor{blue!10}80\% & 
    \cellcolor{blue!10}\textbf{12.18} \\ 
    
    \bottomrule
    \end{tabular}
}
\caption{Comparison of WikiText-2 perplexity for LLaMA-2 7B and LLaMA-3.1 8B models using Duo-SVD and GPTQ.}
\label{tab:quant}
\end{table}

%% file: figures/module_layer_ratio.tex
\begin{figure}[htbp]
    \centering
    
    % --- 第一张子图 ---
    % 关键点：宽度设置为 1.0\linewidth (占满一行)
    \begin{subfigure}{1.0\linewidth}
        \centering
        % 图片本身的宽度可以设小一点，比如 0.6，这样好看
        \includegraphics[width=1\linewidth]{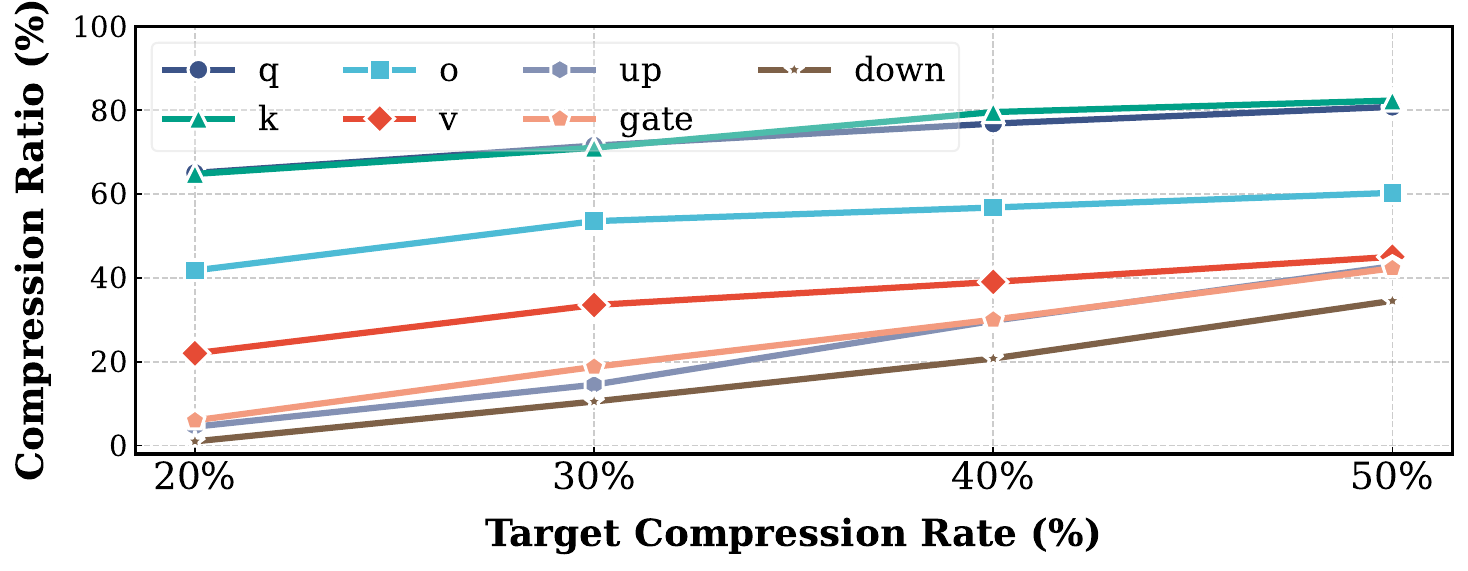}
        \caption{Average compression ratio across different module types}
        \label{fig:module types}
    \end{subfigure}
    
    % 添加垂直间距，防止两图太挤
    \vspace{0.2cm}
    
    % --- 第二张子图 ---
    % 关键点：宽度设置为 1.0\linewidth
    \begin{subfigure}{1.0\linewidth}
        \centering
        \includegraphics[width=1\linewidth]{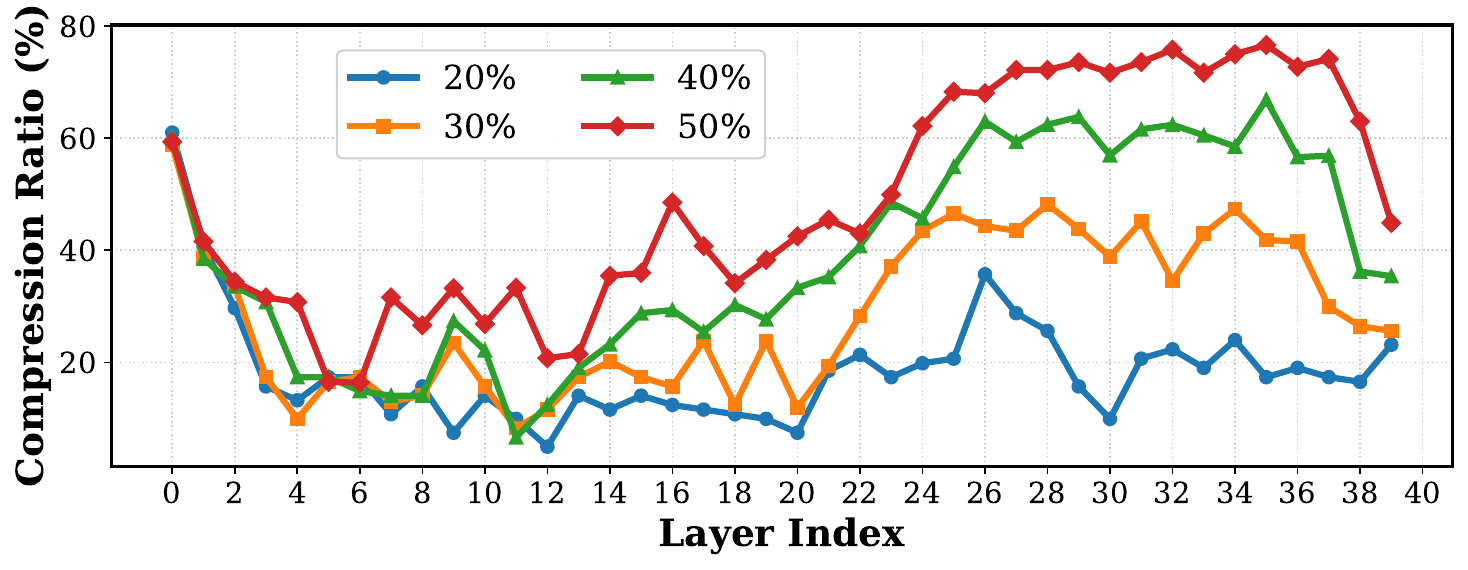}
        \caption{Per-layer compression ratios under different overall compression levels}
        \label{fig:ratio types}
    \end{subfigure}
    
    \caption{Allocation results of LLaMA-2 13B under different compression ratios}
    \label{fig:allocation analysis}
\end{figure}

%% file: tables/inference.tex
\begin{table}[t]
    \centering
    
    \resizebox{\linewidth}{!}{
    \begin{tabular}{cccc}
        \toprule
        \makecell[b]{\textbf{Compression} \\ \textbf{Ratio}} & 
        \makecell[b]{ \textbf{TTFT} \\ (s) $\downarrow$} & 
        \makecell[b]{ \textbf{Throughput} \\ (tokens/s) $\uparrow$} & 
        \makecell[b]{\textbf{Memory} \\ (GB) $\downarrow$} \\
        \midrule
        Dense & 1.3580 & 557.15 & 19.88 \\
        20\%  & 1.2073 & 579.24 & 17.67 \\
        30\%  & 1.1221 & 610.22 & 16.46 \\
        40\%  & 1.0057 & 625.90 & 15.26 \\
        50\%  & 0.9262 & 642.03 & 14.06 \\
        \bottomrule
    \end{tabular}
    }
    \caption{LLaMA-2 7B inference efficiency at different compression ratios. Measured with batch size 24, prefill length 256, and generation length 64.}
    \label{tab:inference}
\end{table}

%% file: figures/maas_ablation.tex
\begin{figure}[htbp]
    \centering
    % width=\linewidth 表示图片宽度等于当前列宽
    % file_name 是您保存的图片文件名，扩展名可以省略
    \includegraphics[width=\linewidth]{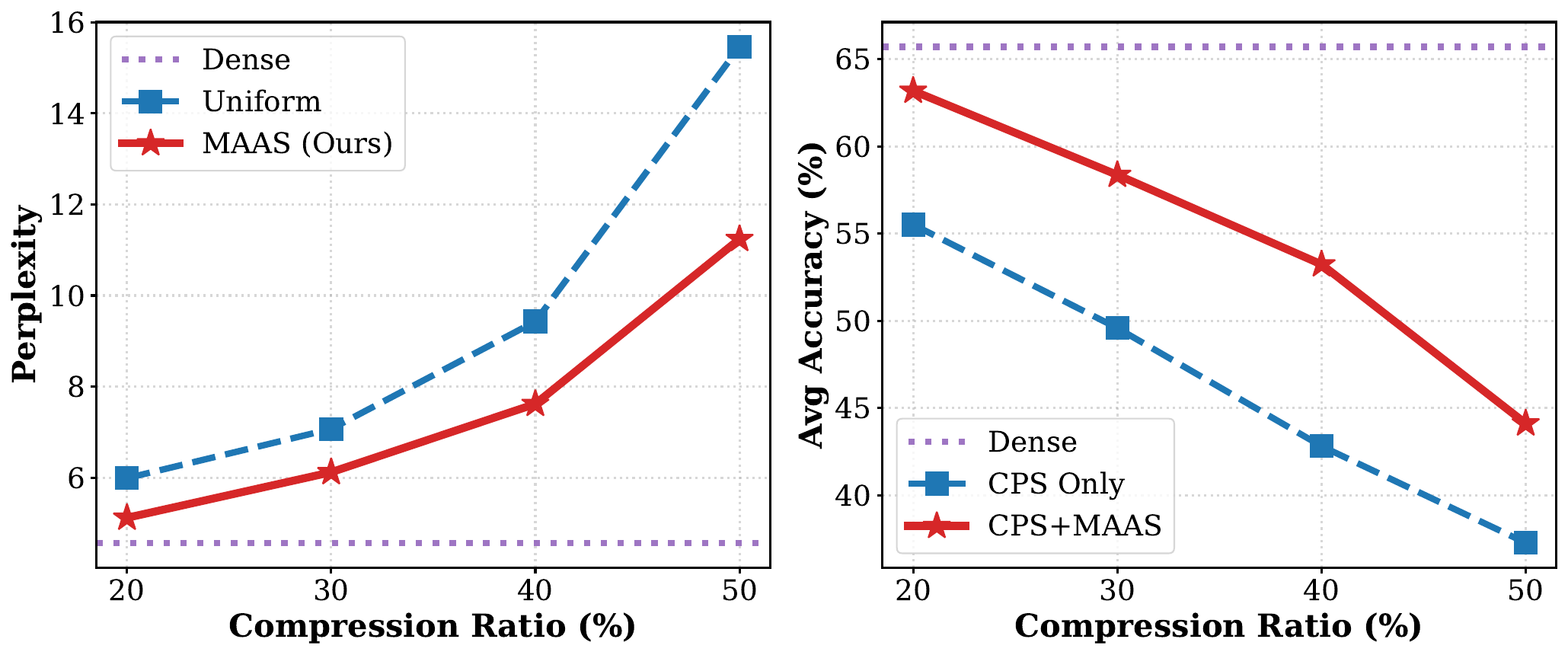}
    
    % 图注
    \caption{Impact of the MAAS on the performance of LLaMA-2 13B.}
    
    % 标签，用于文中引用 (如: see Figure \ref{fig:ppl_comparison})
    \label{fig:maas ablation}
\end{figure}

%% file: figures/calib_datasets.tex
\begin{figure}[htbp]
    \centering
    % width=\linewidth 表示图片宽度等于当前列宽
    % file_name 是您保存的图片文件名，扩展名可以省略
    \includegraphics[width=\linewidth]{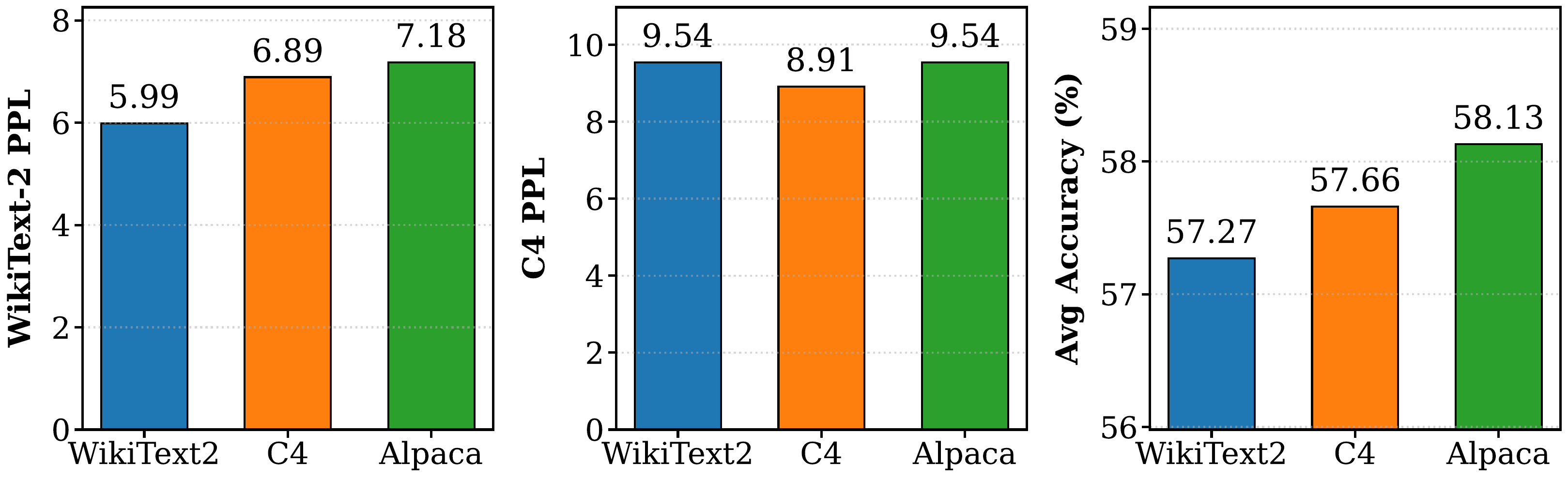}
    
    % 图注
    \caption{Performance comparison of LLaMA-2 7B at 20\% compression ratio using different calibration datasets.}
    
    % 标签，用于文中引用 (如: see Figure \ref{fig:ppl_comparison})
    \label{fig:calibration datasets}
\end{figure}

%% file: components/conclusion.tex
\section{Conclusion}

We proposed \textbf{Duo-SVD}, a training-free framework that optimizes across both column and module levels. Synergizing a \textit{Column-Preserving Strategy} for sensitive weight columns and a \textit{Module-Adaptive Allocation Strategy} for global budget management, Duo-SVD consistently outperforms SOTA baselines across LLaMA and Mistral families. Validated by superior reasoning performance, quantization compatibility, and inference speedups, Duo-SVD establishes a highly effective paradigm for efficient LLM deployment.

%% file: components/Limitations.tex
\section{Limitations}

Although Duo-SVD demonstrates significant reductions in perplexity and theoretical computational costs (FLOPs), a primary limitation lies in the lack of integration within high-performance inference frameworks such as vLLM. Specifically, the proposed \textit{Column-Preserving Strategy} introduces a hybrid computation pattern involving both dense and low-rank pathways. Without specialized hardware-aware optimizations (e.g., fused CUDA kernels), the overhead of memory access and kernel launching in standard implementations may diminish the translation of theoretical compression ratios into real-world wall-clock speedups. Future work will focus on developing optimized kernels to fully realize the acceleration potential of Duo-SVD in production environments. Furthermore, regarding the theoretical analysis, the convexity of the objective function $\mathcal{F}(c)$ is derived based on an approximated upper bound of the reconstruction error. While extensive empirical observations consistently validate the unimodality of the exact loss landscape across diverse settings, a rigorous mathematical proof establishing the strict convexity of the exact $\mathcal{F}(c)$ remains an open theoretical question.

\section{Ethical Considerations}
This work promotes Green AI by reducing resource demands, thereby lowering the barriers to LLM deployment. However, it is important to acknowledge that low-rank decomposition and pruning techniques may inadvertently impact the model's safety alignment or amplify existing biases found in the pre-trained weights. Although our evaluations on standard benchmarks indicate robust performance retention, the potential for degradation in safety guardrails or fairness remains a critical consideration.

%% file: components/appendix.tex
\section{Pseudocode for Duo-SVD}
\label{appendix:pseudocode}

\begin{algorithm}[htbp]
\caption{Duo-SVD Compression Framework}
\label{alg:duo_svd}
\begin{algorithmic}[1]
\REQUIRE Pre-trained Model $\mathcal{W}=\{\mathbf{W}_1, \dots, \mathbf{W}_M\}$, Calibration Data $\mathcal{X}$, Global Target Rate $R_{\text{target}}$, Candidate Rate Set $\mathcal{R}$.
\ENSURE Compressed Model $\mathcal{W}'$.

\STATE \textit{\textbf{// Stage 1: Module-Adaptive Allocation}}
\FOR{each module $i \in \{1, \dots, M\}$}
    \FOR{each rate $s \in \mathcal{R}$}
        \STATE Compute sensitivity $E_{i,s}$ via Eq.~\eqref{eq:module_error} on $\mathcal{X}$.
    \ENDFOR
\ENDFOR
\STATE Solve optimal ratios $\mathbf{s}^* = [s^{(1)}, \dots, s^{(M)}]$ via Dynamic Programming subject to Eq.~\eqref{eq:opt_module}.

\STATE \textit{\textbf{// Stage 2: Column-Preserving Compression}}
\FOR{each module $i \in \{1, \dots, M\}$}
    \IF{$s^{(i)} > 0$}
        \STATE Compute column errors $\{L_{\text{col},j}\}_{j=1}^n$ via Eq.~\eqref{eq:column loss}.
        \STATE Sort errors such that $L_{\text{col},(1)} \ge L_{\text{col},(2)} \ge \dots \ge L_{\text{col},(n)}$.
        \STATE Determine $c^* = \arg\min_c \mathcal{F}(c)$ using \textbf{Ternary Search} based on the convexity of $\mathcal{F}(c)$.
        \STATE Set $\mathcal{S} \leftarrow \{(1), \dots, (c^*)\}$ and calculate rank $r(c^*)$.
        \STATE Decompose $\mathbf{W}_i$ into $\mathbf{W}_{\mathcal{S}}, \mathbf{W}_u, \mathbf{W}_v$ via Eq.~\eqref{eq:hybrid_forward}.
        \STATE $\mathbf{W}'_i \leftarrow \{\mathbf{W}_{\mathcal{S}}, \mathbf{W}_u, \mathbf{W}_v\}$.
    \ELSE
        \STATE $\mathbf{W}'_i \leftarrow \mathbf{W}_i$
    \ENDIF
\ENDFOR
\RETURN $\mathcal{W}' = \{\mathbf{W}'_1, \dots, \mathbf{W}'_M\}$
\end{algorithmic}
\end{algorithm}

\section{Empirical Verification of Convexity}
\label{sec:appendix empirical verification}
% 图二给出了一个函数凸性的验证。这里我们进一步实际检验重构损失和保留列比例之间的凸性。图？？给出了模型不同的层、不同的模型以及不同压缩率下的结果。可以看出，虽然列保留策略在不同设置在的结果差异明显，但他们都展示出了存在唯一极小值点、同时极小值就是最小值的特征。这里的结果充分说明了公式7的近似的合理性，为使用三分搜索快速确定最优的保留列数量提供了坚实的保障。
\input{figures/appendix_loss_r}
\input{tables/appendix_zero_shot_results}

\input{tables/appendix_maas}
Following the preliminary analysis in Figure ~\ref{fig:f vesus c}, we further empirically investigate the convexity of the reconstruction loss with respect to the column retention ratio. Figure~\ref{fig:appendix loss_r} presents the loss landscapes across diverse settings, including different model layers, architectures, and compression ratios. 

Observations indicate that while the specific loss magnitudes vary significantly across these settings, the curves consistently exhibit a unimodal characteristic—possessing a unique local minimum that coincides with the global minimum. These empirical findings strongly validate the approximation proposed in Eq.~\eqref{eq:delta_F_analysis}, providing a solid empirical basis for employing ternary search to efficiently determine the optimal number of retained columns.

\section{Derivation of Convexity for \texorpdfstring{$m \ge n-c$}{m >= n-c}}
\label{sec:appendix_proof}

In the main text, we derived the gradient approximation under the assumption $m \le n-c$. Here, we provide the derivation for the complementary case where $m \ge n-c$. Let $\mathbf{W}_c \in \mathbb{R}^{m \times (n-c)}$ and $\mathbf{W}_{c+k} \in \mathbb{R}^{m \times (n-c-k)}$. Since $m \ge n-c$, the summation of singular values extends only up to the number of columns.
The discrete gradient is derived as follows:
\begin{equation}
\begin{split}
&\Delta \mathcal{F}(c) \\
&= L_{\text{SVD}, r(c)-1}^2(\mathbf{W}_{c+k}) - L_{\text{SVD}, r(c)}^2(\mathbf{W}_{c}) \\
&= \sum_{i=r(c)}^{n-c-k}\sigma_{i}^2(\mathbf{W}_{c+k}) - \sum_{i=r(c)+1}^{n-c}\sigma_{i}^2(\mathbf{W}_{c}) \\
&= \sigma_{r(c)}^2(\mathbf{W}_{c+k}) \\
&\quad \underbrace{+ \sum_{i=r(c)+1}^{n-c-k} \left( \sigma_{i}^2(\mathbf{W}_{c+k}) - \sigma_{i}^2(\mathbf{W}_{c}) \right)}_{\text{Negative Term 1}} \\
&\quad \underbrace{- \sum_{i=n-c-k+1}^{n-c}\sigma_{i}^2(\mathbf{W}_{c})}_{\text{Negative Term 2}} \\
&\approx \underbrace{\sigma_{r(c)}^2(\mathbf{W}_{c+k})}_{\le \delta_{r(c)}^2(\mathbf{W})} - \sum_{j=1}^{k} L_{\text{col}, (c+j)}^2 \\
&\le \delta_{r(c)}^2(\mathbf{W}) - \sum_{j=1}^{k} L_{\text{col}, (c+j)}^2.
\end{split}
\end{equation}

Similar to the main text, the \textit{Negative Term} (consisting of the Cauchy interlacing difference and the tail singular values) represents the total loss of spectral energy due to removing $k$ columns. We approximate this term using the loss of the newly preserved columns ($-\sum L_{\text{col}}^2$).
Consequently, we arrive at the same inequality structure: the upper bound is composed of a increasing cost term $\delta_{r(c)}^2(\mathbf{W})$ and a decreasing gain term $\sum L_{\text{col}}^2$. Thus, the convexity of $\mathcal{F}(c)$ holds regardless of the matrix dimensions.

\section{Additional Comparison of Different Methods}
\label{sec:appendix additional zero shot results}

% 这一节我们给出Duo-SVD和更多的baseline在零样本任务上的比较，其中，LaCo,ShotGPT和BlockPruner是深度剪枝方法，剪枝的基本单元是transformer层或者attention和mlp层；Basis Sharing和Dobi-SVD基于SVD分解。对于Dobi-SVD，为了公平比较，我们使用其未结合量化方法的版本。具体实验结果如表？？？所示，Duo-SVD在平均精度上超过基线方法至少4个百分点，展示出优秀的性能保留效果。

In this section, we present a comparative analysis of Duo-SVD against an expanded set of baselines on zero-shot tasks. Specifically, LaCo, ShortGPT, and BlockPruner represent depth pruning approaches that operate at the granularity of Transformer layers or individual Attention and MLP modules. In contrast, Basis Sharing and Dobi-SVD are founded on SVD decomposition. To ensure a fair comparison, we evaluate the unquantized version of Dobi-SVD. As reported in Table~\ref{tab:appendix zero shot results}, Duo-SVD outperforms the baseline methods by a margin of at least 4 percentage points in average accuracy, demonstrating superior performance retention capabilities.

\section{Additional Results of MAAS}
\label{sec:appendix maas}
\input{figures/appendix_maas}

% 这里我们给出LLaMA-2 7B，Mistral 7B， LLaMA-3 8B在20%压缩率下的不同类型的模块的压缩率以及不同层的压缩率的MAAS结果。如图？？？所示，不同模型展现出相似性，他们的attn模块的压缩率明显比mlp模块更高，后半部分的压缩率比前半部分更高。同时模型之间也有区别：LLaMA-3 8B模型的v模块压缩更多的比例同时k模块保留更多，Mistral 7B模型相对而言中间层的压缩率更高。这里的结果对分析各个模型的模块重要性和层重要性提供了新的数据。
In this section, we present the MAAS results regarding the compression rates of distinct module types and layers for LLaMA-2 7B, Mistral-7B, and LLaMA-3.1 8B, all evaluated at a 20\% compression ratio. 
As illustrated in Figure~\ref{fig:appendix allocation analysis}, the models exhibit consistent patterns: the compression ratios for Attention modules are significantly higher than those for MLP modules, and the latter layers are compressed more heavily than the earlier ones. 
However, distinct model-specific behaviors are also observable. Specifically, in LLaMA-3.1 8B, the $V$ modules undergo more aggressive compression, whereas the $K$ modules retain a larger proportion of parameters. In contrast, Mistral-7B exhibits relatively higher compression rates in its intermediate layers. 
Collectively, these findings provide novel data for analyzing the relative importance of specific modules and layers across different model architectures.

\section{Comparison of Different Allocation Strategy}
\label{sec: appendix different allocation strategy}
% 为了检验MAAS的效果，这里我们将其和其他的压缩率分配方法进行比较。具体来说，我们使用Bolaco中使用的贝叶斯优化和SoLA使用的最小化各模块截断误差的和作为baseline。表？？给出的llama 2 7b在20%和30%压缩率下的结果显示我们提出的MAAS在困惑度和下游任务的精度上均能够取得更好的结果，展示出和贝叶斯优化为了降低优化维度而使用的多个模块贡献同一个秩，以及SoLA通过截断误差这一间接指标，我们的MAAS通过对直接衡量各个模块的性能能够获得更好的表现。
To evaluate the effectiveness of MAAS, we compare it against distinct compression ratio allocation strategies. For clarity in Table~\ref{tab:appendix ratio methods comp}, we denote the Bayesian optimization framework employed by Bolaco~\cite{bolaco} as BayesOpt, and the cumulative truncation error minimization approach adopted by SoLA~\cite{sola} as TEM. To ensure a fair comparison, we uniformly employ CPS as the underlying decomposition technique across all allocation methods.

The results for LLaMA2-7B at 20\% and 30\% compression ratios, presented in Table~\ref{tab:appendix ratio methods comp}, demonstrate that MAAS consistently achieves superior performance in both perplexity and downstream task accuracy. The primary advantage of MAAS stems from its capability to directly quantify the contribution of each module to the model's overall performance. Unlike BayesOpt, which necessitates coarse-grained rank sharing to handle optimization dimensionality, or TEM, which relies on truncation error as an indirect proxy, MAAS enables a fine-grained and precise rank allocation. This direct evaluation mechanism allows MAAS to effectively preserve critical information, thereby yielding the significant performance gains observed over the baselines.

\section{Theoretical Analysis of Efficiency}
\label{sec:appendix efficiency analysis}

Based on the dual-pathway formulation defined in Eq.~\ref{eq:hybrid_forward} and the optimization constraint in Eq.~\ref{eq:opt_column}, we analyze the theoretical efficiency of Duo-SVD.
\paragraph{Storage Efficiency.}
The storage efficiency is directly enforced by the optimization constraint. For a target compression ratio $\rho \in (0, 1)$, the total parameter budget is strictly bounded by $N_{\text{Duo}} \le \rho mn$. Assuming the budget is fully utilized, the storage reduction ratio is trivially $N_{\text{Duo}} / N_{\text{orig}} = \rho$.

\paragraph{Computational Efficiency.}
To prove the linear reduction in computation, we express the total parameter count $N_{\text{Duo}}$ as the sum of the dense and low-rank components:
\begin{equation}
\label{eq:param_identity}
    N_{\text{Duo}} = mc + r(m + n - c) = \rho mn.
\end{equation}
The total computational cost $\mathcal{C}_{\text{Duo}}$ (in FLOPs) for a batch size $d$ aggregates the operations from both paths:
\begin{equation}
    \mathcal{C}_{\text{Duo}} = \underbrace{d \cdot mc}_{\text{Dense Path}} + \underbrace{d \cdot r(n - c) + d \cdot mr}_{\text{Low-Rank Path}}.
\end{equation}
By factoring out $d$ and rearranging the terms, we observe that the bracketed expression is identical to the parameter definition in Eq.~\ref{eq:param_identity}:
\begin{equation}
\begin{split}
    \mathcal{C}_{\text{Duo}} &= d \cdot \left[ mc + r(m + n - c) \right] \\
    &= d \cdot N_{\text{Duo}} \\
    &= d \cdot \rho mn.
\end{split}
\end{equation}
Since the original computational cost is $\mathcal{C}_{\text{orig}} = d \cdot mn$, the reduction ratio is exactly $\mathcal{C}_{\text{Duo}} / \mathcal{C}_{\text{orig}} = \rho$. This confirms that both storage and computation scale linearly with the compression ratio $\rho$.
% 添加对于凸函数假设的更加广泛的验证：不同的模型，不同的层，不同的压缩率
% 添加对于MAAS的有效性实验，和Bolaco，SoLA的比较

%% file: figures/appendix_loss_r.tex
\begin{figure*}[htbp]
    \centering
    
    % --- 第一行 ---
    \begin{subfigure}[b]{0.32\linewidth}
        \centering
        \includegraphics[width=\linewidth]{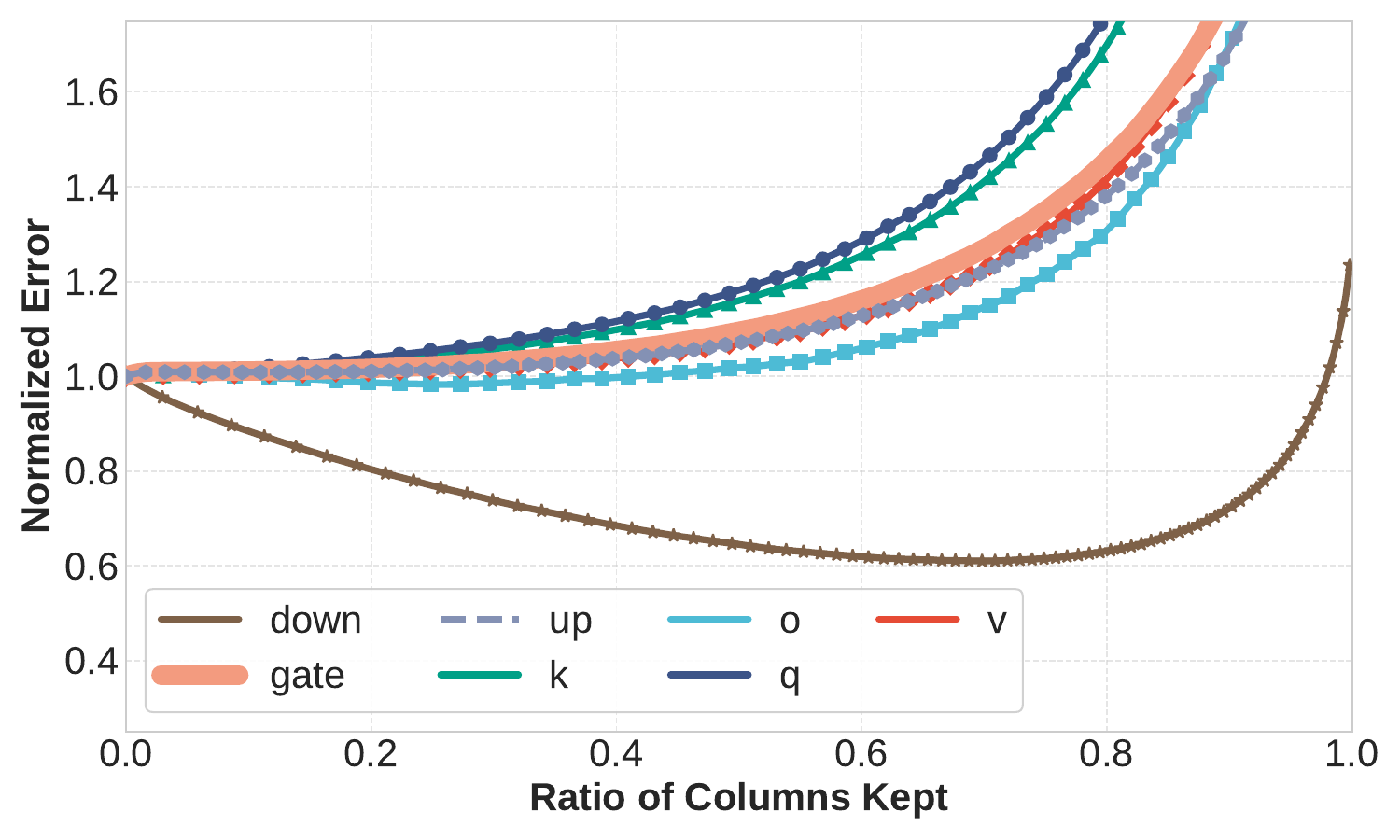} % 换成你的图片文件名
        \caption{Layer 11}
        \label{fig:layer 11}
    \end{subfigure}
    \hfill % 图片之间的弹性间距
    \begin{subfigure}[b]{0.32\linewidth}
        \centering
        \includegraphics[width=\linewidth]{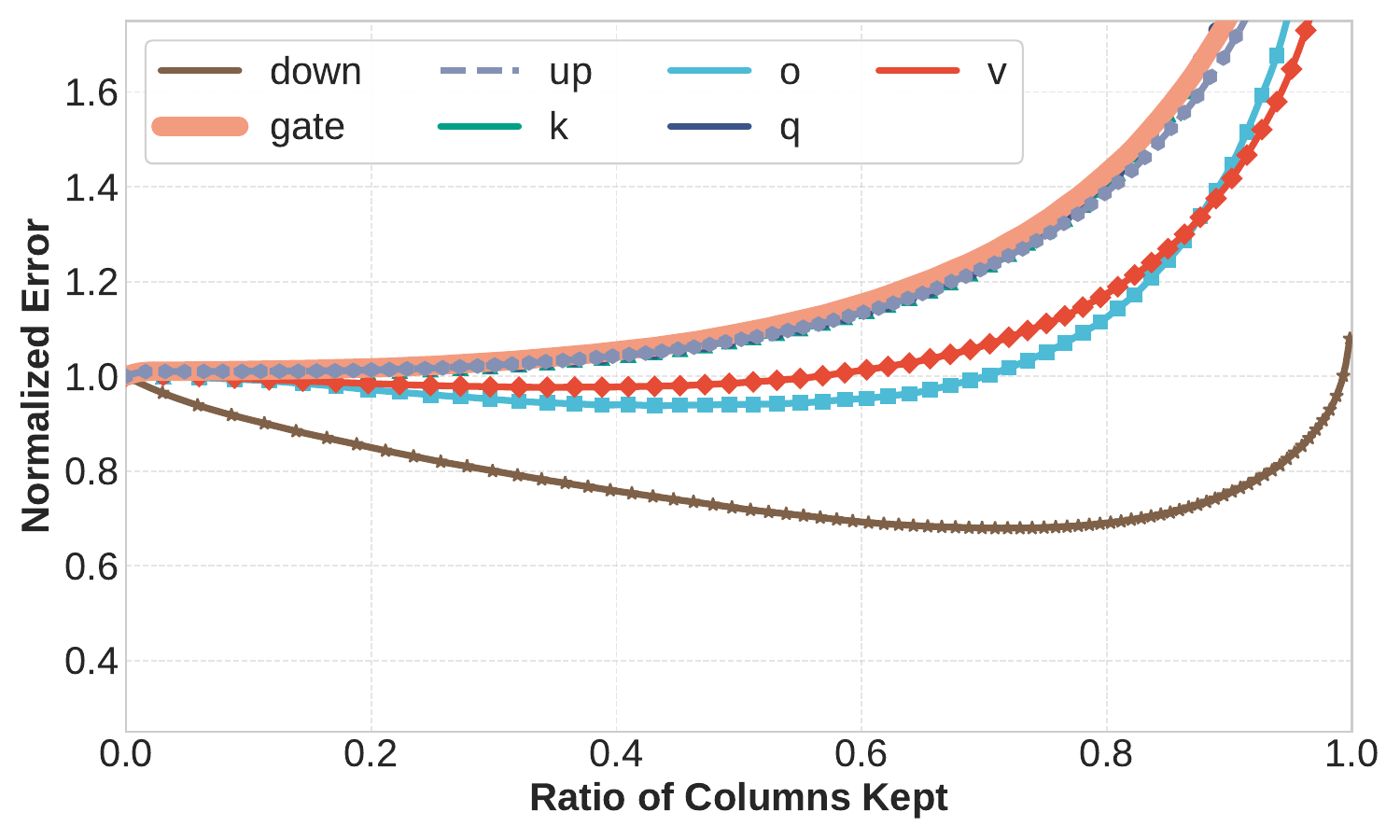}
        \caption{Layer 21}
        \label{fig:layer 21}
    \end{subfigure}
    \hfill
    \begin{subfigure}[b]{0.32\linewidth}
        \centering
        \includegraphics[width=\linewidth]{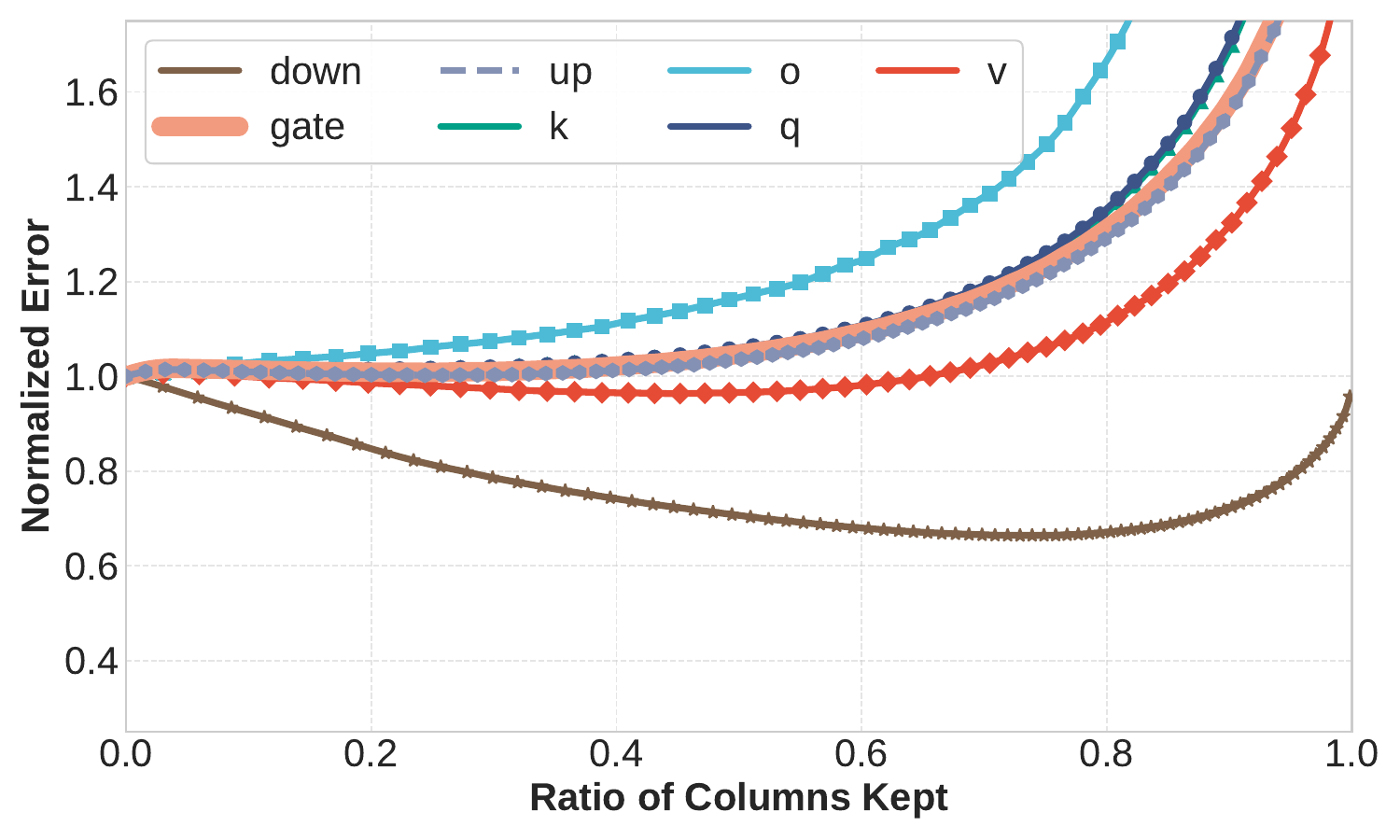}
        \caption{Layer 31}
        \label{fig:layer 31}
    \end{subfigure}
    
    \vspace{0.5cm} % 行间距
    
    % --- 第二行 ---
    \begin{subfigure}[b]{0.32\linewidth}
        \centering
        \includegraphics[width=\linewidth]{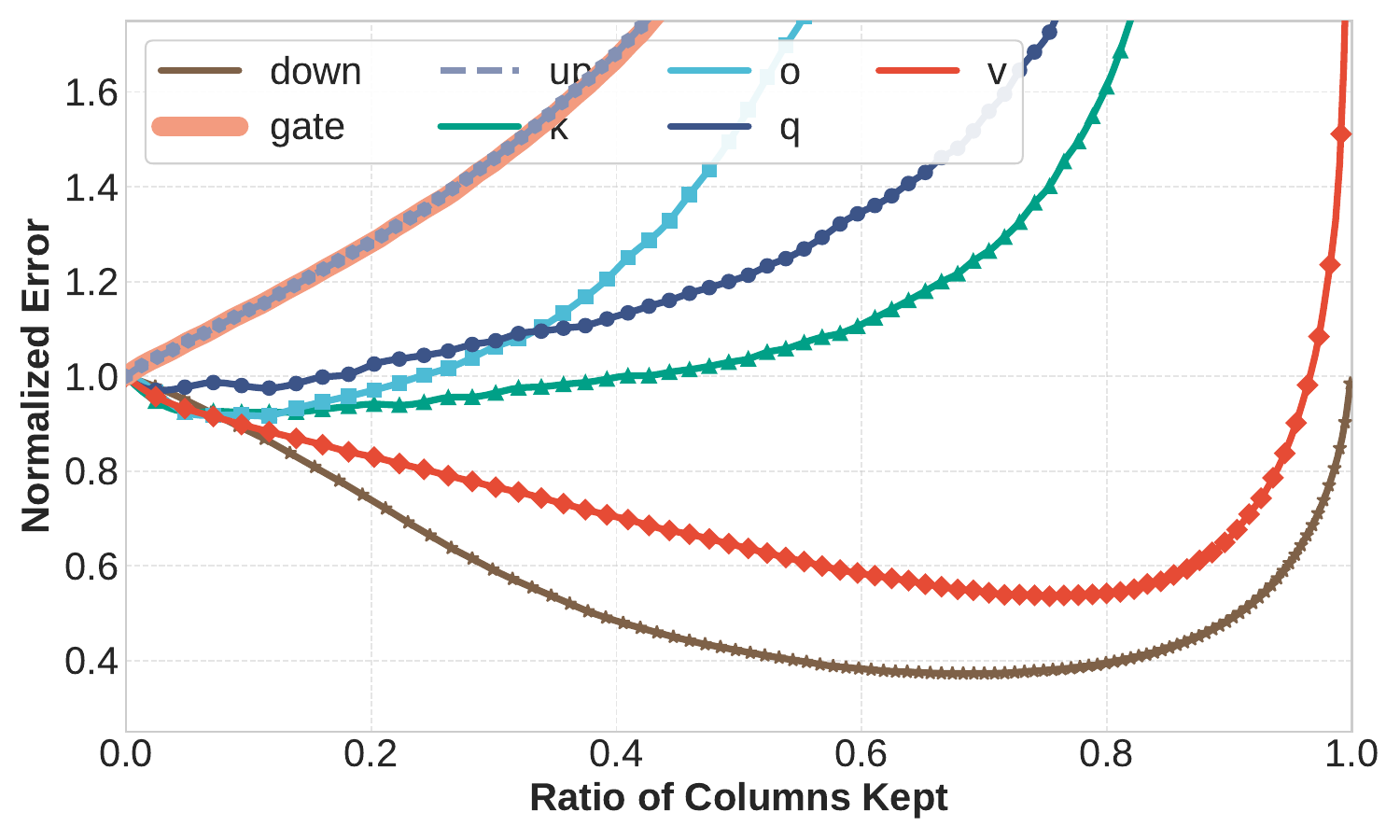}
        \caption{LLaMA-2 13B}
    \end{subfigure}
    \hfill
    \begin{subfigure}[b]{0.32\linewidth}
        \centering
        \includegraphics[width=\linewidth]{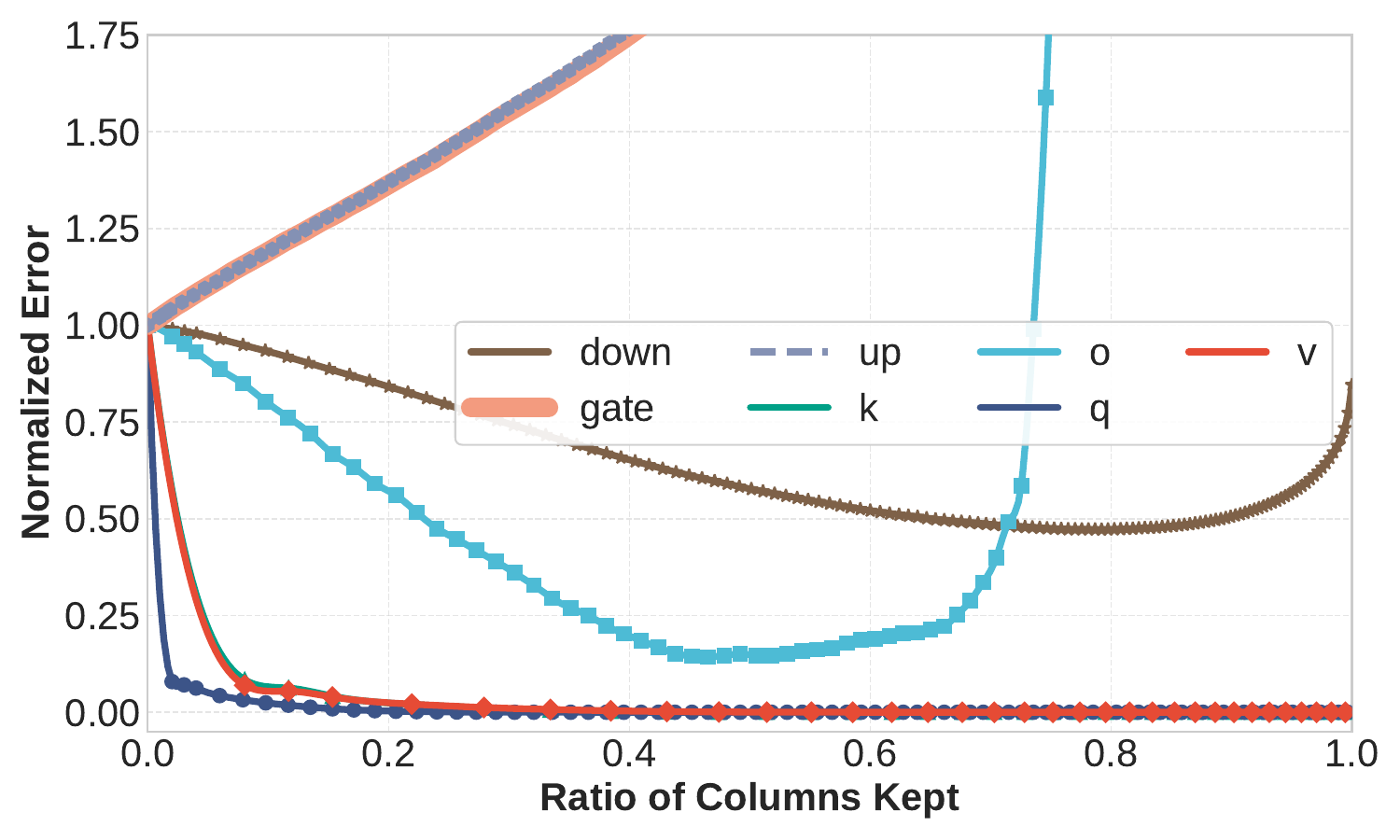}
        \caption{Mistral-7B}
    \end{subfigure}
    \hfill
    \begin{subfigure}[b]{0.32\linewidth}
        \centering
        \includegraphics[width=\linewidth]{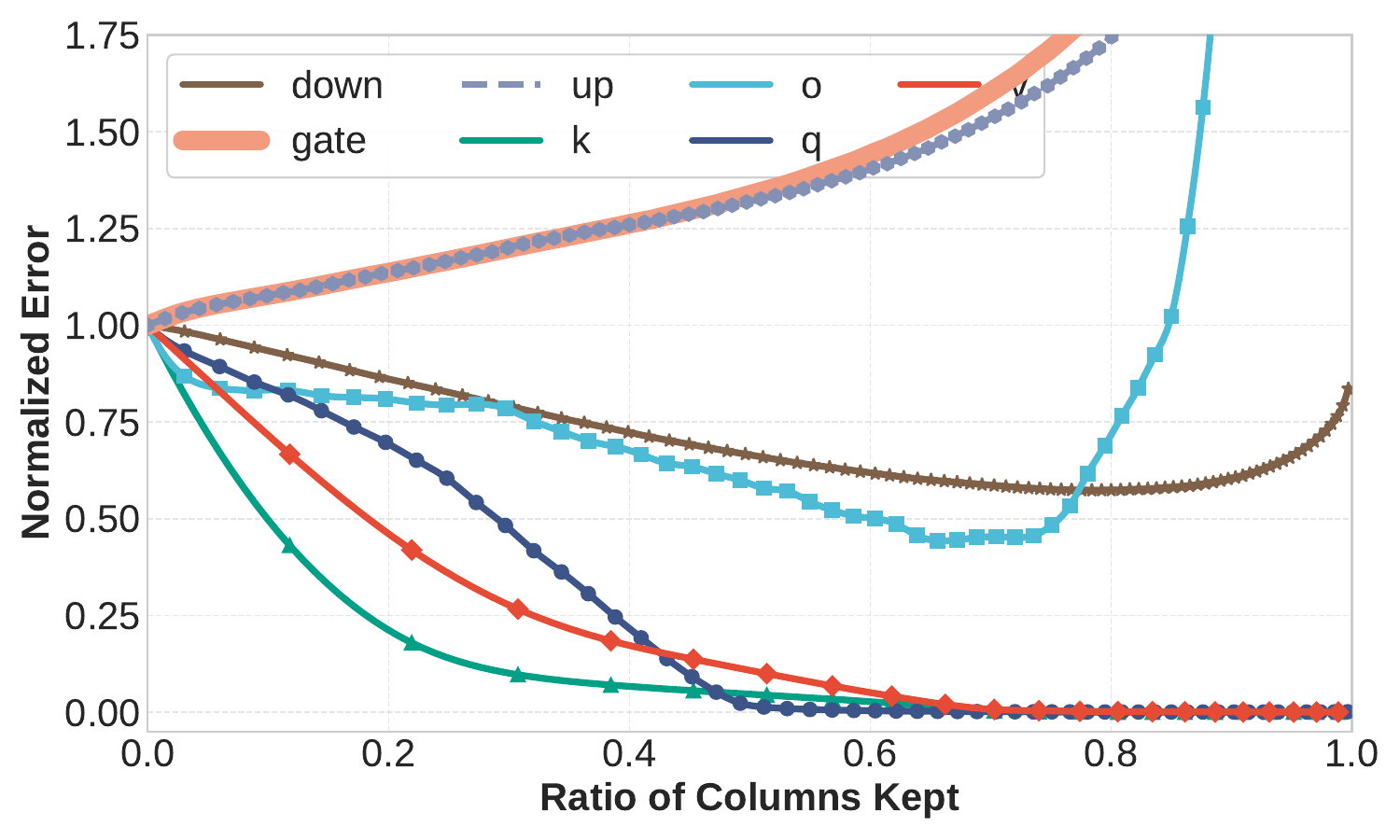}
        \caption{LLaMA-3.1 8B}
    \end{subfigure}
    
    \vspace{0.5cm} % 行间距
    
    % --- 第三行 ---
    \begin{subfigure}[b]{0.32\linewidth}
        \centering
        \includegraphics[width=\linewidth]{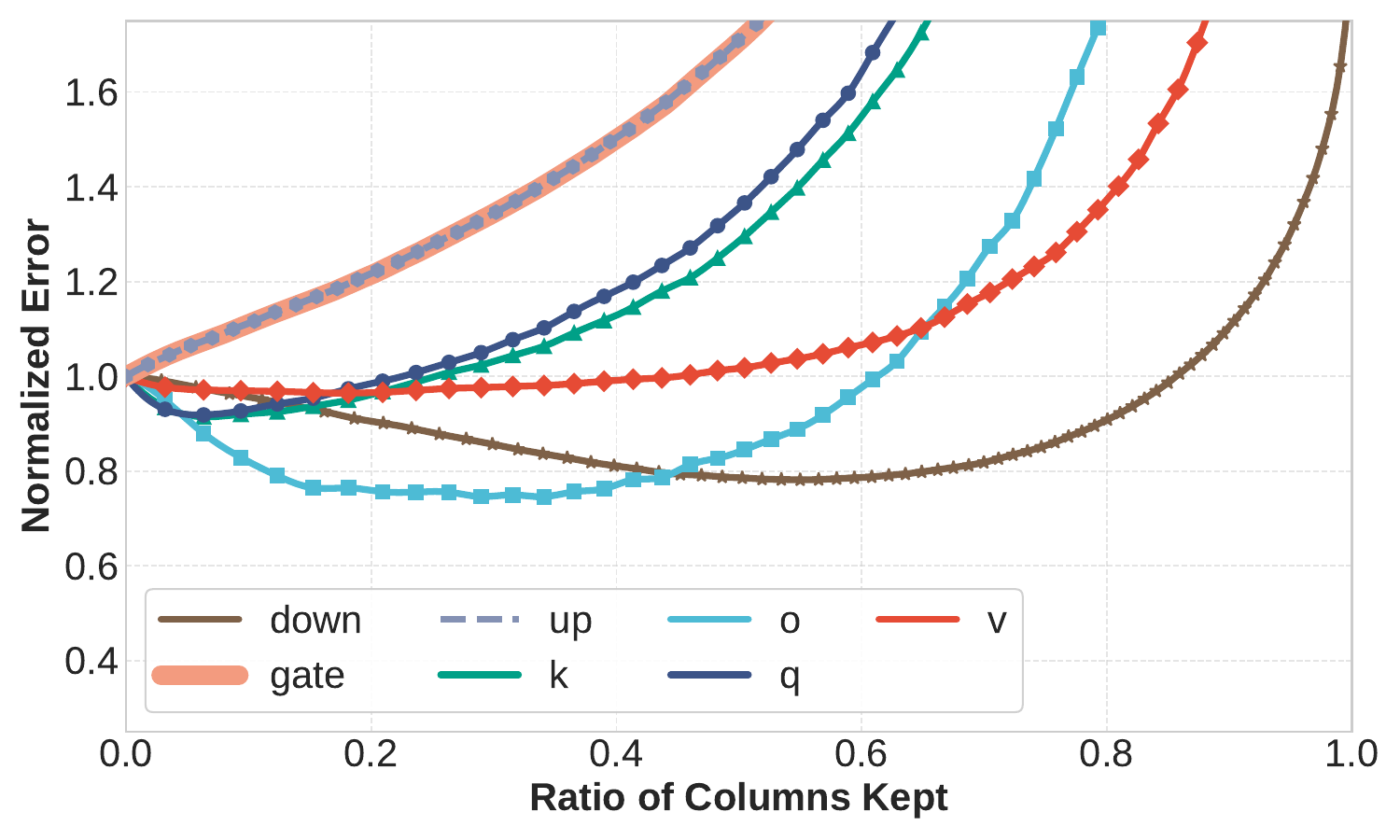}
        \caption{30\%}
    \end{subfigure}
    \hfill
    \begin{subfigure}[b]{0.32\linewidth}
        \centering
        \includegraphics[width=\linewidth]{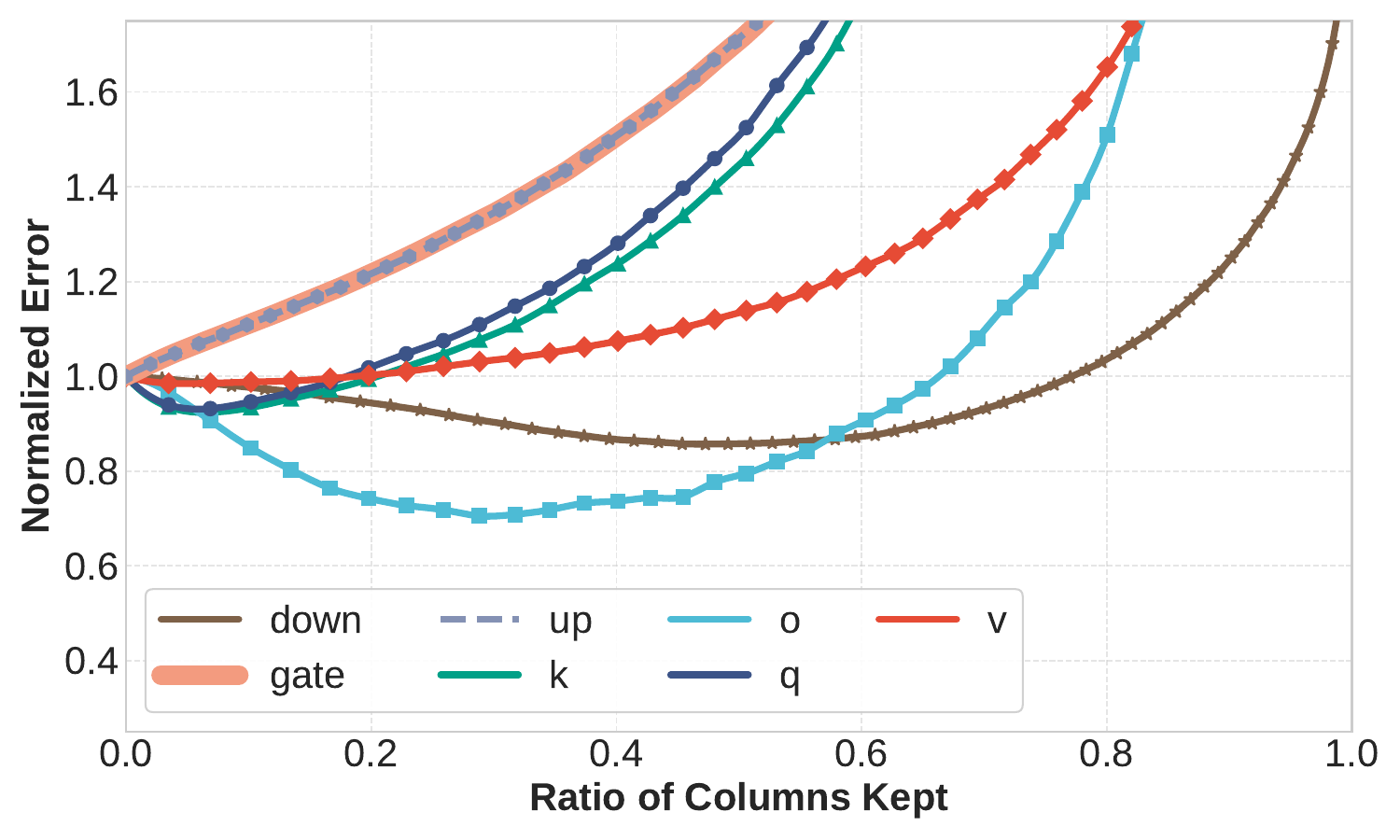}
        \caption{40\%}
    \end{subfigure}
    \hfill
    \begin{subfigure}[b]{0.32\linewidth}
        \centering
        \includegraphics[width=\linewidth]{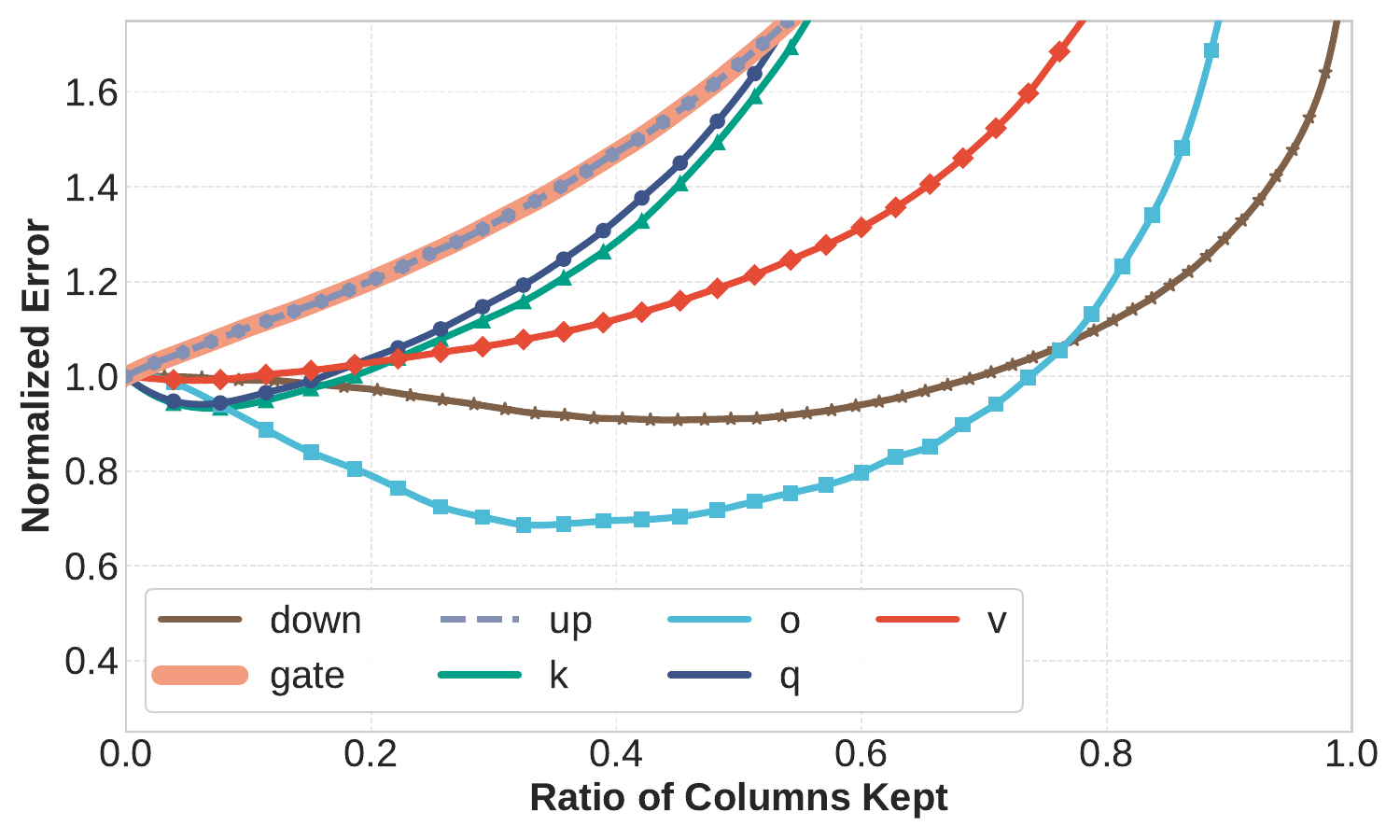}
        \caption{50\%}
    \end{subfigure}
    
    \caption{Verification of the monotonicity of $\mathcal{F}(c)$. 
    (a)--(c) Results for LLaMA-2 7B at a 20\% compression ratio. 
    (d)--(f) Results for Layer 0 at a 20\% compression ratio. 
    (g)--(i) Results for Layer 0 of LLaMA-2 7B.}
    \label{fig:appendix loss_r}
\end{figure*}

%% file: tables/appendix_zero_shot_results.tex
\begin{table*}[htbp]
\centering
\resizebox{\textwidth}{!}{%
\begin{tabular}{c|c|ccccc|c}
\toprule
\textbf{Model} & \textbf{Method} & \textbf{WinoG.} & \textbf{HellaS.} & \textbf{ARC-e} & \textbf{ARC-c} & \textbf{PIQA} & \textbf{Avg.} $\uparrow$ \\ 
\midrule
\multirow{7}{*}{\textbf{LLaMA-2 7B}} 
 & Dense & 69.06 & 75.99 & 74.58 & 46.25 & 77.91 & 68.76 \\
 \midrule
 & LaCo & 60.46 & 54.08 & 55.39 & 35.84 & 68.34 & 54.82 \\
 & ShortGPT & 65.90 & 62.63 & 56.06 & 36.09 & 70.24 & 58.18 \\
 & BlockPruner & 62.43 & 65.87 & 61.07 & 37.29 & 74.21 & 60.17 \\
 & Basis Sharing & \textbf{66.54} & 56.22 & 59.18 & 30.82 & 66.54 & 55.86 \\
 & Dobi-SVD & 58.56 & 56.80 & 54.25 & 26.29 & 65.34 & 52.25 \\
 \rowcolor{blue!10} \cellcolor{white} & \textbf{Duo-SVD} & 65.67 & \textbf{70.23} & \textbf{70.50} & \textbf{40.53} & \textbf{76.77} & \textbf{64.74} \\
\bottomrule
\end{tabular}%
}
\caption{Performance comparison of different methods on LLaMA-2 7B at 20\% compression ratio}
\label{tab:appendix zero shot results}
\end{table*}

%% file: tables/appendix_maas.tex
\begin{table*}[htbp]
\centering
\setlength{\tabcolsep}{3pt}
% \resizebox{\textwidth}{!}{% 自动调整表格宽度
\begin{tabular}{c|c|c|c|ccccccc}
\hline
\textbf{Ratio} & \textbf{Method} & \textbf{PPL} $\downarrow$ & \textbf{Avg.} $\uparrow$ & \textbf{MMLU-5shot} & \textbf{PIQA} & \textbf{WinoG.} & \textbf{HellaS.} & \textbf{ARC-e} & \textbf{ARC-c} & \textbf{OBQA} \\ \hline
Dense & -- & 5.11 & 62.14 & 45.70 & 79.05 & 69.38 & 75.92 & 74.49 & 46.25 & 44.20 \\ \hline 
 & BayesOpt & 6.50 & 55.24 & 33.84 & 73.99 & \textbf{65.98} & 64.08 & 69.19 & 37.97 & 41.60 \\
 & TEM & 8.05 & 46.26 & 28.68 & 70.95 & 63.77 & 55.98 & 60.48 & 34.13 & 38.80 \\
\rowcolor{blue!10} \cellcolor{white}\multirow{-3}{*}{\textbf{20\%}} & \textbf{MAAS} & \textbf{5.99} & \textbf{57.27} & \textbf{35.41} & \textbf{76.77} & 65.67 & \textbf{70.23} & \textbf{70.50} & \textbf{40.53} & \textbf{41.80} \\ \hline 
 & BayesOpt & 7.85 & 50.45 & \textbf{29.87} & 70.18 & \textbf{65.11} & 55.87 & 59.97 & 33.36 & \textbf{38.80} \\
 & TEM & 11.60 & 44.12 & 29.68 & 63.28 & 60.22 & 44.28 & 49.07 & 28.50 & 33.80 \\
\rowcolor{blue!10} \cellcolor{white}\multirow{-3}{*}{\textbf{30\%}} & \textbf{MAAS} & \textbf{7.50} & \textbf{51.11} & 27.09 & \textbf{72.14} & 63.69 & \textbf{60.97} & \textbf{61.28} & \textbf{34.13} & \textbf{38.80} \\ \hline 

\end{tabular}
% }
\caption{Comparison of WikiText-2 perplexity and downstream task accuracy of different methods on LLaMA-2 7B.}
\label{tab:appendix ratio methods comp}
\end{table*}

%% file: figures/appendix_maas.tex
\begin{figure}[htbp]
    \centering
    
    % --- 第一张子图 ---
    % 关键点：宽度设置为 1.0\linewidth (占满一行)
    \begin{subfigure}{1.0\linewidth}
        \centering
        % 图片本身的宽度可以设小一点，比如 0.6，这样好看
        \includegraphics[width=1\linewidth]{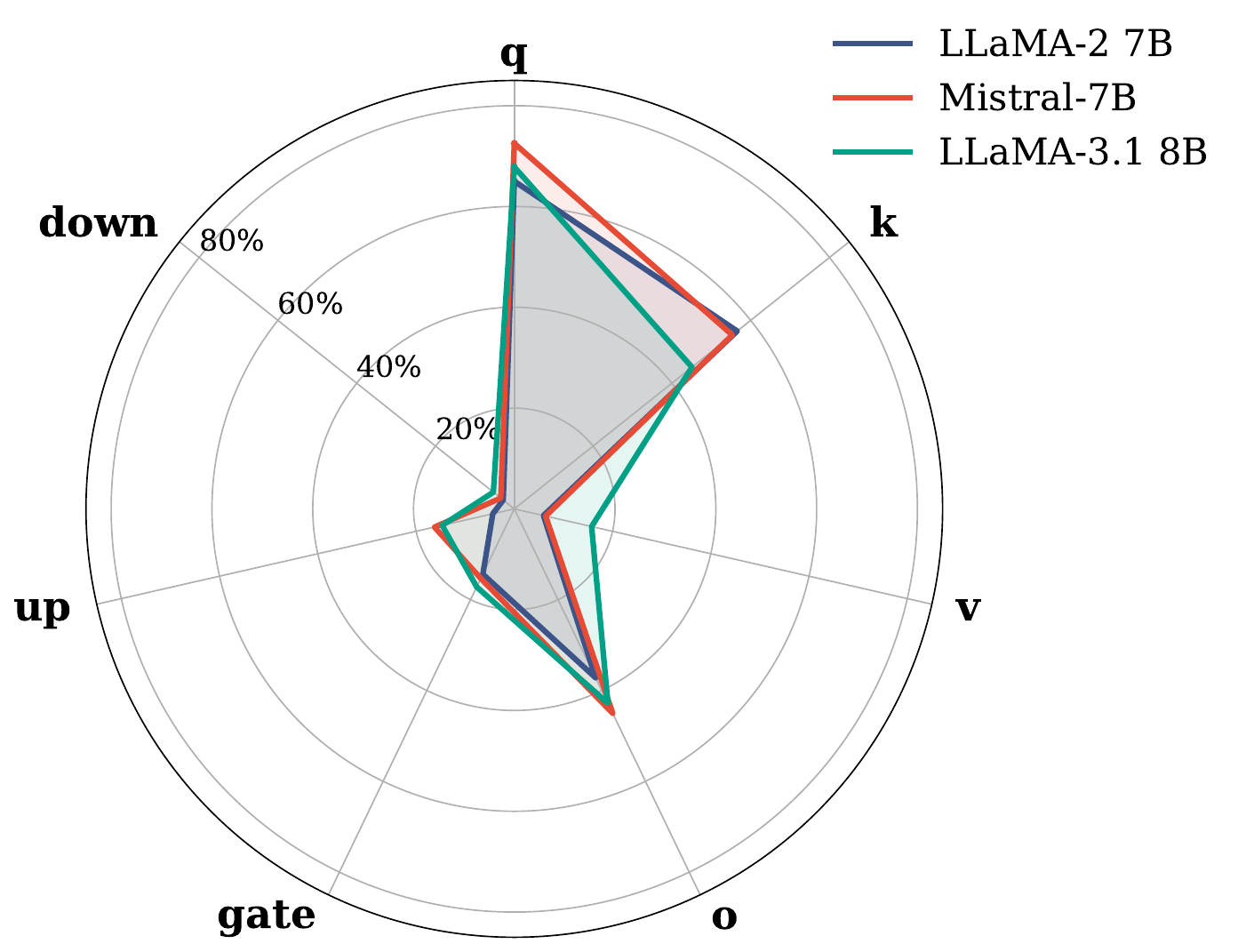}
        \caption{Average compression ratio across different module types}
        \label{fig:appendix model module ratio}
    \end{subfigure}
    
    % 添加垂直间距，防止两图太挤
    \vspace{0.2cm}
    
    % --- 第二张子图 ---
    % 关键点：宽度设置为 1.0\linewidth
    \begin{subfigure}{1.0\linewidth}
        \centering
        \includegraphics[width=1\linewidth]{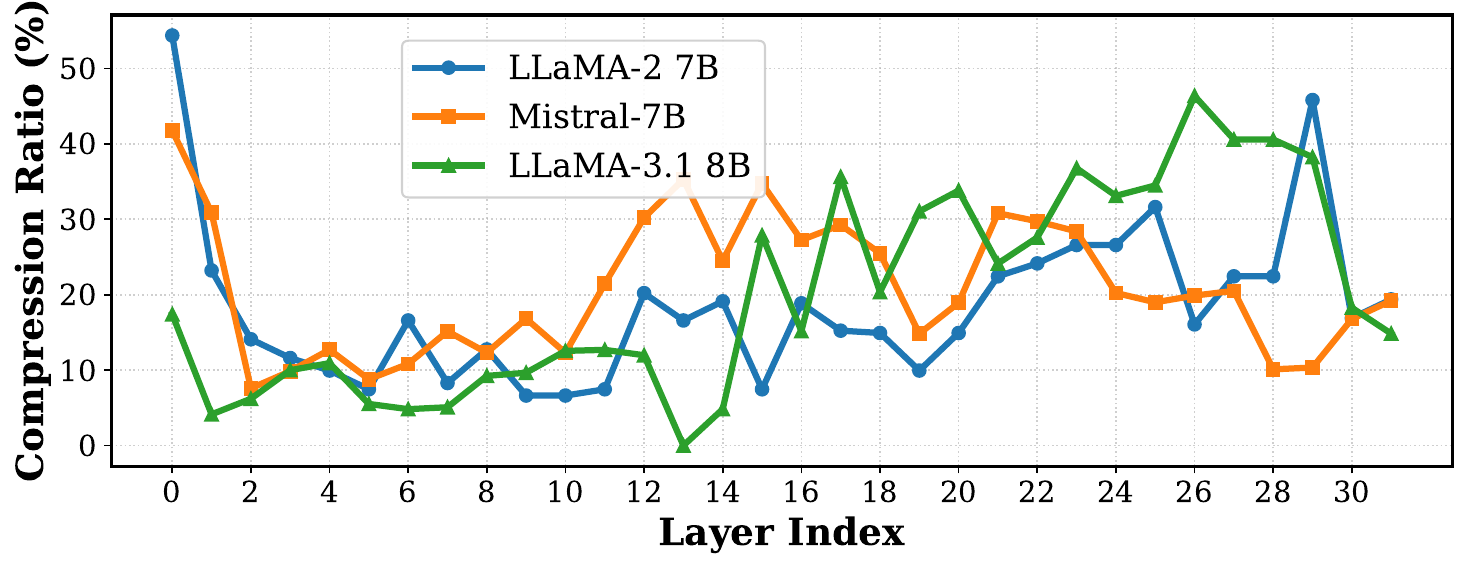}
        \caption{Per-layer compression ratios}
        \label{fig:appendix model layer ratio}
    \end{subfigure}
    
    \caption{Allocation results of different models at 20\% compression ratio}
    \label{fig:appendix allocation analysis}
\end{figure}